\pdfoutput=1

\documentclass[11pt,dvipsnames]{article}

\usepackage[final]{acl}

\usepackage{amsmath,amsfonts,amssymb,bm}

\def\eqref#1{equation~\ref{#1}}

\def\1{\bm{1}}

\DeclareMathAlphabet{\mathsfit}{\encodingdefault}{\sfdefault}{m}{sl}
\SetMathAlphabet{\mathsfit}{bold}{\encodingdefault}{\sfdefault}{bx}{n}

\newcommand{\softmax}{\mathrm{softmax}}

\usepackage{times}
\usepackage{latexsym}

\usepackage[T1]{fontenc}

\usepackage[utf8]{inputenc}

\usepackage{microtype}

\usepackage{inconsolata}

\usepackage{hyperref}       \usepackage{url}            \usepackage{booktabs}       \usepackage{amsfonts}       \usepackage{nicefrac}       \usepackage{xcolor}         

\usepackage{graphicx}
\usepackage{subcaption}
\usepackage{tabularx}
\newcolumntype{Y}{>{\centering\arraybackslash}X}
\usepackage{numprint}
\usepackage{multirow}
\npthousandsep{,}\npthousandthpartsep{}\npdecimalsign{.}
\usepackage{color, colortbl}
\definecolor{lightgrey}{gray}{0.9}
\definecolor{lightblue}{rgb}{0.68, 0.85, 0.9}
\definecolor{lightcyan}{rgb}{0.88, 1.0, 1.0}
\usepackage{enumitem}
\usepackage{xspace} 
\newcommand{\method}{\textsc{TokenTune}\xspace}
\usepackage{stmaryrd}
\usepackage{cleveref}
\usepackage{numprint}
\usepackage{makecell}
\usepackage{titlesec}
\titlespacing{\paragraph}{0pt}{2ex}{0.1cm}
\usepackage[ruled,vlined,linesnumbered]{algorithm2e}
\usepackage{setspace}

\newcommand\blfootnote[1]{\begingroup
    \renewcommand\thefootnote{}\footnote{#1}\addtocounter{footnote}{-1}\endgroup
}

\newcommand{\githubURL}{\url{https://github.com/facebookresearch/tokentune}}

\title{Memory-Efficient Fine-Tuning of Transformers via Token Selection}

\author{
\textbf{Antoine Simoulin\textsuperscript{\textbf{*}}},
\textbf{Namyong Park\textsuperscript{\textbf{*}}},
\textbf{Xiaoyi Liu},
\textbf{Grey Yang}
\\
Meta AI
\\
\small{
	\texttt{\{antoinesimoulin,namyongp,xiaoyiliu,glyang\}@meta.com}
}
}

\begin{document}
\maketitle

\begin{abstract}
Fine-tuning provides an effective means to specialize pre-trained models for various downstream tasks. However, fine-tuning often incurs high memory overhead, especially for large transformer-based models, such as LLMs. While existing methods may reduce certain parts of the memory required for fine-tuning, they still require caching all intermediate activations computed in the forward pass to update weights during the backward pass. In~this work, we develop \method, a method to reduce memory usage,  specifically the memory to store intermediate activations, in the fine-tuning of transformer-based models. During the backward pass, \method approximates the gradient computation by backpropagating through just a subset of input tokens. Thus, with \method, only a subset of intermediate activations are cached during the forward pass. Also, \method can be easily combined with existing methods like LoRA, further reducing the memory cost. We evaluate our approach on pre-trained transformer models with up to billions of parameters, considering the performance on multiple downstream tasks such as text classification and question answering in a few-shot learning setup. Overall, \method achieves performance on par with full fine-tuning or representative memory-efficient fine-tuning methods,  while greatly reducing the memory footprint, especially when combined with other methods with complementary memory reduction mechanisms. We hope that our approach will facilitate the fine-tuning of large transformers,  in specializing them for specific domains or co-training them with other neural components from a larger system. Our code is available at \githubURL.
\blfootnote{\textbf{*} Equal contribution}
\end{abstract}

\section{Introduction}
\label{introduction}

Fine-tuning is an effective method for specializing large pre-trained models, either by using direct supervision from the training set of a given task \citep{ruder_18, devlin_19, raffel_20}, from curated instruction datasets \citep{DBLP:conf/acl/MishraKBH22, wei_22, alpaca}, or from human feedback via reinforcement learning \citep{ouyang_22, bai_22, touvron_23}. 
However, fine-tuning is not necessarily an efficient method,
especially for transformer-based large language models (LLMs),
since their large number of parameters leads to large compute and memory requirements.
For instance, fine-tuning GPT-3 175B~\citep{DBLP:conf/nips/BrownMRSKDNSSAA20} or LLama 65B~\citep{touvron_23} 
typically requires 1,200 GB and 780 GB of GPU memory, as reported in \citet{hu_22} and \citet{dettmers_23}, respectively.

 \begin{figure}[t]
\centering
 	\makebox[0.4\textwidth][c]{
 		\includegraphics[width=1.02\columnwidth]{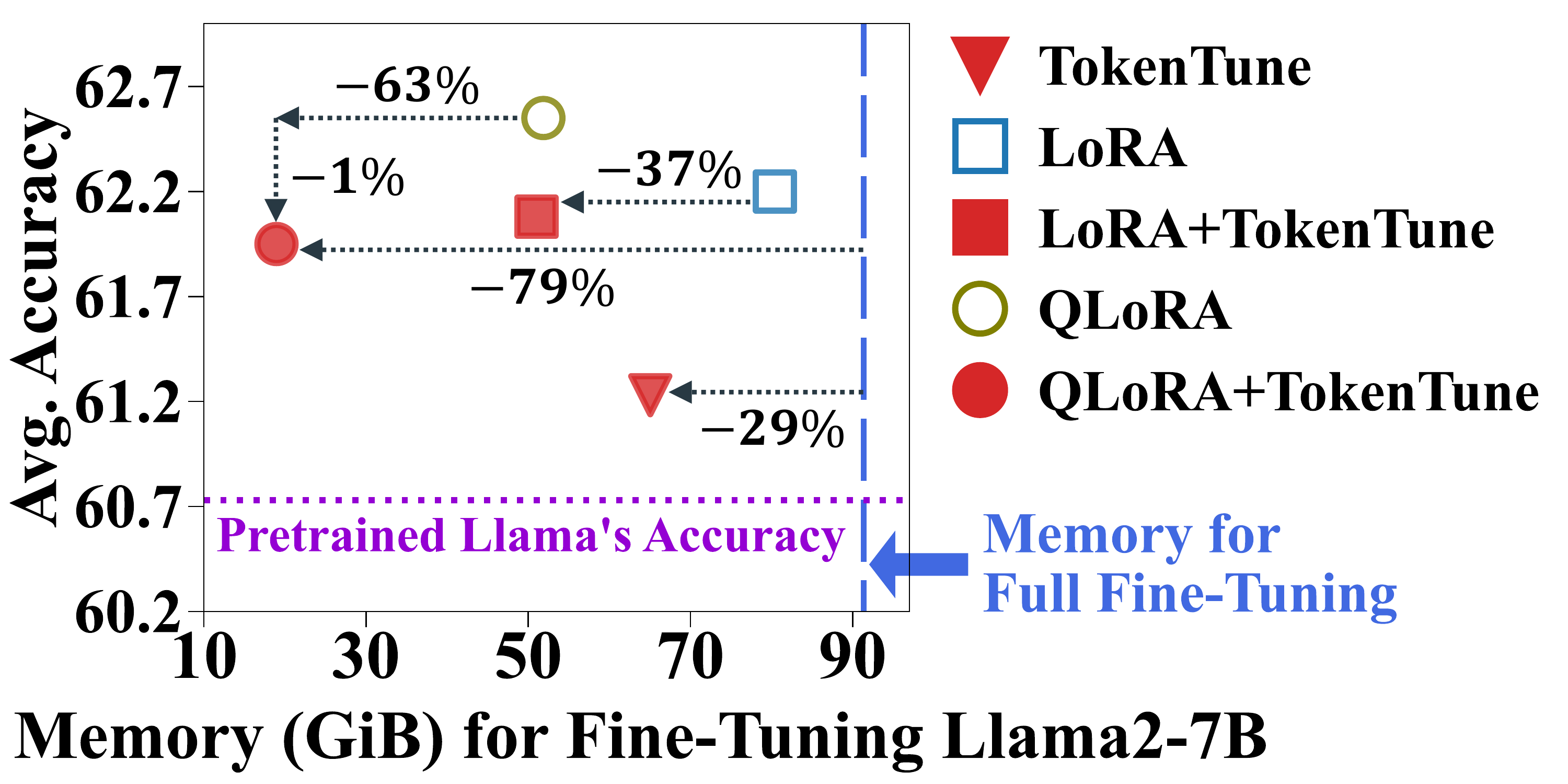}
 	}
\caption{\method greatly \mbox{reduces} the GPU memory usage for fine-tuning the \mbox{Llama2-7B} model 
(e.g., using only 37\% of the memory~QLoRA~\citep{dettmers_23} requires),
while achieving similar accuracy to representative memory-efficient fine-tuning methods.
 		Accuracy and memory usage numbers are listed in \Cref{tab:llm-perf} and Fig.~\ref{fig:llm-memory}.
 		See Sec.~\ref{sec:exp:large} for details on experiments.
 	}
 	\label{fig:crownjewel}
 \end{figure}

GPU memory usage during fine-tuning can be broken down into three parts: 
storing (1) the model parameters, (2) the parameter gradients and optimizer states, and (3) the intermediate activations. 
Parameter-Efficient Fine-Tuning (PEFT)~\citep{DBLP:conf/icml/HoulsbyGJMLGAG19, hu_22} aims at updating a small number of parameters, 
e.g., by optimizing a subset of the backbone model's parameters while freezing others,
which reduces the memory requirements to store the parameters' gradients and optimizer states.
Alternatively, quantization techniques~\citep{dettmers_22, dettmers_23, liu_23} use low precision data types for model parameters, which reduces the memory cost.
For example, in fine-tuning the Llama2-7B model,
LoRA~\citep{hu_22} and QLoRA~\citep{dettmers_23}, which are representative PEFT and quantization-based methods,
reduce the memory needed for full fine-tuning by 12\% and 43\%, respectively (\Cref{fig:crownjewel}).
However, such existing approaches still require caching all of the intermediate activations computed in the forward pass
to obtain the gradients during the backward pass.

In this work, we propose a method for memory-efficient fine-tuning, named \method, 
which aims to significantly reduce the GPU memory dedicated to storing intermediate activations during the forward pass
without sacrificing the model performance on various downstream tasks.
To this end, \method selects a subset of the input tokens in the context, and fine-tunes the model with respect to those selected tokens.
More specifically, during the backward pass, \method approximates the gradient computation by backpropagating through the selected tokens, and thus 
only a subset of the intermediate activations need to be cached during the forward pass, thereby reducing the memory cost.

We demonstrate the effectiveness of \method using both medium- and large-size language models, namely, BERT~\citep{devlin_19} and Llama~\citep{touvron_23}, 
which have hundreds of millions, and billions of parameters, respectively.
Overall, our results show that fine-tuning with \method leads to downstream task performance on par with that of full fine-tuning or representative methods for memory-efficient fine-tuning,
while drastically reducing the memory footprint.
Notably, \method can be effectively combined with existing methods, achieving a greater reduction in memory usage.
For instance, by combining \method with QLoRA~\citep{dettmers_23}, 
we can fine-tune Llama2-7B using just about one third of the memory QLoRA alone requires as \Cref{fig:crownjewel} shows.
To~sum, our contributions are as follows.
\begin{itemize}[leftmargin=1em,topsep=-2pt,itemsep=-3pt]
	\item \textbf{Novelty.} \method, to the best of our knowledge, is the first method that reduces GPU memory usage for fine-tuning via token selection\footnote{A preliminary version of this work was presented at a non-archival workshop~\citep{simoulin2023memoryefficient}.}.
	\item \textbf{Combinability.} \method can be combined with existing memory-efficient fine-tuning methods, leading to further memory reduction.
	\item \textbf{Effectiveness.} We perform extensive experiments, showing that 
	\method achieves similar accuracy to representative memory-efficient methods,
	while greatly reducing the memory footprint during fine-tuning,
	e.g., using only 21\% of what full fine-tuning requires (\Cref{fig:crownjewel}).
\end{itemize}

\section{Related Work}

\label{sec:related-work}
\begin{figure*}[t]
\centering
	\makebox[0.4\textwidth][c]{
		\includegraphics[width=1.0\textwidth]{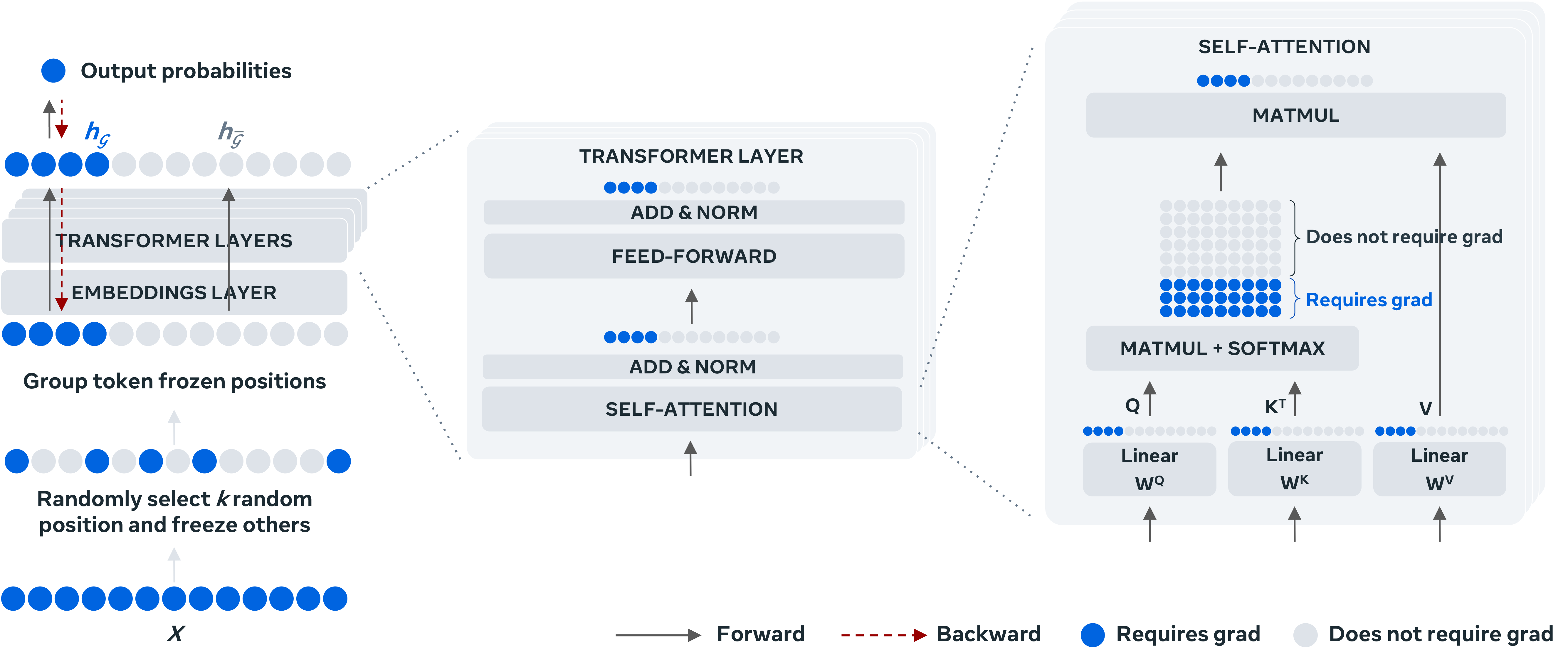}
}
\caption{\method achieves memory-efficient fine-tuning of transformers via token selection.
		During the backward pass, we compute the gradient for only a subset of $k$ input tokens, while the others are frozen (in \textcolor{gray}{gray} in the figure).
		During the forward pass, all input positions are used, but only a subset of the activations is cached~in~memory (in \textcolor{blue}{blue} in the figure).
		\method is applicable to various transformer-based models, as well as different language modeling tasks, as our experiments with \textsc{Bert}~\citep{devlin_19} and Llama~\citep{touvron_23} show.
}
\label{fig:method}
\end{figure*}

\subsection{Parameter-Efficient Fine-Tuning (PEFT)}
PEFT methods, which aim to limit the computing resources for fine-tuning LLMs, 
can be divided into four categories~\citep{DBLP:journals/corr/abs-2403-14608,DBLP:journals/corr/abs-2312-12148}.

\paragraph{Selective PEFT} methods update only a subset of the backbone model parameters
using weight masking strategies, such as
learnable binary masking~\citep{DBLP:conf/acl/GuoRK20} and parameter importance estimation using Fisher information~\citep{DBLP:conf/nips/SungNR21,DBLP:conf/emnlp/DasZS0Z23}. Other selective PEFT methods focus on updating specific modules,
e.g., the cross-attention layers~\citep{DBLP:conf/emnlp/Gheini0M21} and the bias terms~\citep{zaken_22,DBLP:conf/acl/LawtonKTGS23}.  

\paragraph{Additive PEFT} methods add a few parameters to the frozen pre-trained model, 
and fine-tune only the added parameters. E.g., adapters inject small layers within the transformer block,
either sequentially after its sublayers~\citep{DBLP:conf/icml/HoulsbyGJMLGAG19,DBLP:conf/eacl/PfeifferKRCG21}, or as a side network running in parallel to the sublayers~\citep{DBLP:conf/iclr/HeZMBN22,DBLP:conf/emnlp/ZhuFZWL21}.
Alternatively, soft prompt-based approaches~\citep{DBLP:conf/acl/LiL20,qin_21,liu_21} 
prepend continuous learnable vectors to the input of a frozen model and tune them for each~task.

\paragraph{Reparameterized PEFT} methods perform low-rank transformation,
utilizing the low intrinsic dimension of LLMs~\citep{DBLP:conf/acl/AghajanyanGZ20}.  LoRA~\citep{hu_22} is the most representative approach, 
where an update to the model weights is captured via its low-rank decomposition.
Several studies followed to improve LoRA, e.g., 
to support dynamic rank selection~\citep{DBLP:conf/eacl/ValipourRKG23,DBLP:conf/iclr/ZhangCBH0CZ23}, and to address overfitting~\citep{DBLP:journals/corr/abs-2404-09610} and overconfidence~\citep{DBLP:conf/iclr/YangRWA24}.  

\paragraph{Hybrid PEFT} methods aim to combine different PEFT approaches, e.g.,
adapters, prefix-tuning, and LoRA.
The design space of combinations of PEFT methods has been explored
either manually~\citep{DBLP:conf/iclr/HeZMBN22,DBLP:conf/acl/MaoMHAM0YK22}, or automatically, e.g., by leveraging neural architecture search methods~\citep{DBLP:journals/corr/abs-2206-04673,DBLP:journals/tacl/ZhouWVK24}. 

\paragraph{\kern-0.3em}
While the above PEFT methods effectively improve parameter efficiency,
they may still incur significant memory overhead during fine-tuning~\citep{sung2022lst,DBLP:conf/emnlp/JinZZ23}.
The proposed \method can be combined with these PEFT methods,
enabling them to achieve both parameter and memory efficiency,
as \Cref{sec:exp:medium,sec:exp:large} show.

\subsection{Memory-Efficient Fine-Tuning}

There exist several techniques that can be used to improve the memory efficiency in fine-tuning LLMs,
which we organize into four groups.

\paragraph{Memory-Efficient PEFT.}
Some PEFT methods aim to achieve memory and parameter efficiency simultaneously.
Side tuning methods~\citep{zhang_20, sung2022lst}
introduce small learnable side networks separated from the backbone model, and
channel backpropagation only through the side networks, 
thereby reducing the memory requirements for gradients and intermediate activations.
By utilizing the reversible model, 
MEFT~\cite{DBLP:conf/nips/LiaoTM23} avoids the need to cache intermediate activations in the forward pass.
LoRA-FA~\citep{DBLP:journals/corr/abs-2308-03303} improves LoRA
by addressing its high memory usage for input activations
via freezing LoRA's down-projection weights.

\paragraph{Gradient Checkpointing}~\citep{DBLP:journals/corr/ChenXZG16,DBLP:conf/nips/GruslysMDLG16}
reduces the memory requirement for model training
by storing only a subset of intermediate activations in the forward pass, and recomputing the others during the backward pass.

\paragraph{Quantization} is a compression technique that reduces the number of bits for storing numerical values. 
With quantization, parameters are represented with lower-precision data types~\citep{dettmers_22, dettmers_23, liu_23},
leading to memory reduction in both fine-tuning and inference.

\paragraph{Approximate Gradient Methods} reduce the memory usage
by avoiding the exact gradient computation involved with full fine-tuning, and 
instead using an approximate estimate of the gradient for weight updates.
To this end, a few methods employ low-rank factorization,
where they reduce memory cost by utilizing 
the low-rank structure of the gradients~\citep{zhao2024galore} or the second-order statistics~\citep{DBLP:conf/icml/ShazeerS18}. Alternatively, MeZO~\citep{DBLP:conf/nips/MalladiGNDL0A23} 
approximates the gradient using only forward passes,
building upon the zeroth-order optimization technique~\cite{Spall1992MultivariateSA}.

\paragraph{\kern-0.3em}
The proposed \method can be considered an approximate gradient method, 
as its token-selective fine-tuning strategy leads to an approximation of the full gradient,
which is a completely new direction investigated to improve memory efficiency in fine-tuning.
Also, being complementary to prior methods, 
\method can be combined with them, resulting in further memory reduction.

\section{\method}  \label{sec:method}

Previous studies analyzing the structure of the sparsity of activations and gradients~\citep{DBLP:conf/icml/KurtzKGMCGLMSA20,DBLP:conf/icml/LiuWDZY0S0TRC23, DBLP:conf/acl/DaiDHSCW22}
suggest that some neurons and activations could have a predominant importance, while some others may have smaller contributions to the loss and output computation.
Inspired by these works, we hypothesize that for many downstream tasks, 
not all tokens in the sequence would need to be involved in the fine-tuning---more specifically, backpropagation---of transformer models.
Instead, we conjecture that, when restricted to backpropagating through a subset of tokens, 
transformers could be further optimized for the downstream task 
by enabling the additional learning and adjustments, which need to happen during the fine-tuning for the given task, 
to be done in a more compact way, i.e., by incorporating the additional knowledge more succinctly with respect to the selected subset of tokens.

Figure~\ref{fig:method} illustrates \method, aiming at reducing the memory needed to store the intermediate activations used for gradient computation. Given an input sequence $X$, a transformer associates each token from the input sequence to an embedding and computes a corresponding sequence of hidden states $h$ through multiple~layer~applications.
For each input sequence, we select $k$ random positions.\footnote{We select the positions using a uniform distribution. However, we always include the \texttt{[CLS]} token---a special symbol prepended as the beginning of every input sentence.} We organize each layer's input in two groups, 
one with the $k$ selected input positions,~$h_{\mathcal{G}}$, and the other with the remaining un-selected positions, $h_{\bar{\mathcal{G}}}$, such that $h = [ h_{\mathcal{G}}, h_{\bar{\mathcal{G}}}]$, with $[\quad]$ denoting the concatenation operator and $\bigm|\mathcal{G}\bigm|=k$. The re-ordering does not impact the computation as the position is directly encoded in the hidden states. 
With this token selection scheme, the classification objective $\mathcal{L}_{\text{CLS}}$ and the language modeling objective $\mathcal{L}_{\text{LM}}$ used by \method are as follows.

\paragraph{Classification Task.} 
The goal is to assign the right class or label $y$ for the given sequence.
Given the hidden states from the transformer layers,
we use the average of the hidden states from the $k$ selected positions of the last layer as input for an MLP,
which outputs a probability distribution over the classes of the task, as given by Eq.~\ref{eq: pooler}. 
During the evaluation, we use the average from all hidden states of the last layer as input for the MLP.
\begin{equation}
\begin{split}
    \pi &= \text{MLP}\left(\frac{1}{k}\sum_{i \in \mathcal{G}}h_{i}\right)\\
    p(y | X) &= \text{softmax}(\pi) \\
    \mathcal{L}_{\text{CLS}} &= -\log p(y | X)
\end{split}
  \label{eq: pooler}
\end{equation}

\paragraph{Language Modeling Task.} 
The goal is to learn the probability distribution of a token, given all preceding tokens.
We train the language model  by applying the traditional cross-entropy loss to the set of $k$ randomly selected positions as given by Eq.~\ref{equ:ref_loss} below, with $W_{\text{lm}}$ denoting the head projecting the hidden state back into the vocabulary dimension.

\begin{equation}
\begin{split}
p(x_i | x_{<i}) &= \text{softmax}(h_{i} W_{\text{lm}} ) \\
\mathcal{L}_{\text{LM}} &= -\sum_{i \in \mathcal{G}} \log P(x_i | x_{<i})
\end{split}
\label{equ:ref_loss}
\end{equation}
The key element of our method is that we disable the gradient computation for the un-selected tokens in $\bar{\mathcal{G}}$. 
Thus, only the $k$ selected tokens in $\mathcal{G}$ contribute to the gradient computation during the backward~pass. We detail the method in the case of dense layers and attention mechanism in \Cref{method:dense} and \Cref{method:att}, respectively.

\subsection{\method for Dense and Normalization Layers}
\label{method:dense}

We consider a dense layer $a = \sigma (z) = \sigma (hW + b)$ with weight $W$, bias $b$, nonlinear function~$\sigma$, input $h$, pre-activation~$z$, and output~$a$.
Eq.~\ref{eq: gradient of weights} computes the gradient with respect to $W$ and $b$ when backpropagating a loss~$\mathcal{L}$ through the layer:
\begin{align}\label{eq: gradient of weights}
\begin{split}
 \frac{\partial \mathcal{L}}{dW} &= \frac{\partial \mathcal{L}}{\partial a}\frac{\partial a}{\partial z}\frac{\partial z}{\partial W} =
    \frac{\partial \mathcal{L}}{\partial a} \sigma' h \\
\frac{\partial \mathcal{L}}{db} &= \frac{\partial \mathcal{L}}{\partial a}\frac{\partial a}{\partial z}\frac{\partial z}{\partial b} = \frac{\partial \mathcal{L}}{\partial a} \sigma'
\end{split}
\end{align}If we backpropagate the error only through the selected tokens in $\mathcal{G}$, and disable the gradient computation for the unselected positions in $\bar{\mathcal{G}}$, we have:
\begin{align}
    \frac{\partial \mathcal{L}}{\partial a} = \left[ \frac{\partial \mathcal{L}}{\partial a_{\mathcal{G}}}, \frac{\partial \mathcal{L}}{\partial a_{\bar{\mathcal{G}}}} \right] = \left[ \frac{\partial \mathcal{L}}{\partial a_{\mathcal{G}}}, 0 \right]
\end{align}

Plugging that into Eq.~\ref{eq: gradient of weights}, we have:
\begin{align} \label{eq: gradient of weights final}
    \frac{\partial \mathcal{L}}{dW} = \left[ \frac{\partial \mathcal{L}}{\partial a_{\mathcal{G}}} \sigma' h_{\mathcal{G}}, 0 \right];
    \hspace{0.5em} \frac{\partial \mathcal{L}}{db} = \left[ \frac{\partial \mathcal{L}}{\partial a_{\mathcal{G}}} \sigma', 0 \right]
\end{align}
Given Eq.~\ref{eq: gradient of weights final}, we only need to cache $h_{\mathcal{G}}$ for applying the chain rule, instead of the full activation~$h$. 

Regarding implementation, we use Algorithm~\ref{alg:tokentune} which explicitly splits the hidden states into two groups where $h_{\mathcal{G}}$ corresponds to the tokens selected to be fine-tuned and $h_{\bar{\mathcal{G}}}$ corresponds to the un-selected tokens. As shown in Eq.~\ref{eq:dense:v} and Eq.~\ref{eq:dense:no-grad}, the forward pass is identical to standard fine-tuning except that we disable the gradient computation for the positions for $h_{\bar{\mathcal{G}}}$ in Eq.~\ref{eq:dense:no-grad} with the context "\texttt{torch.no\_grad()}" in PyTorch. 
\begin{align}
h_{\mathcal{G}} &= h_{\mathcal{G}}W + b \label{eq:dense:v}\\
h_{\bar{\mathcal{G}}} &= h_{\bar{\mathcal{G}}}W + b\label{eq:dense:no-grad}
\end{align}
where $W$ denotes the weights $W_1$ and $W_2$ for the feed-forward layers. We apply the same methodology for normalization layers.

\begin{table*}[t]
    \centering
\caption{Results from \textsc{Bert}-large \citep{devlin_19} on GLUE test tasks scored using the benchmark server. We report the Matthew’s Correlation for CoLA, the Spearman correlation for STS-B, F1 score for MRPC and QQP. We report the accuracy on the MNLI matched test split and the accuracy for every other tasks. The ``Param.'' column indicates the ratio of the number of updated parameters for each task by the number of parameters in the backbone model. We indicate in \textbf{bold} the best result for each task. $^{\dagger}$ indicates models we trained. We report adapter results from \citep{DBLP:conf/icml/HoulsbyGJMLGAG19}, BitFit from \citep{zaken_22} and Diff Pruning from \citep{DBLP:conf/acl/GuoRK20}. For LoRA \citep{hu_22} and Ladder Side Tuning (LST) \citep{sung2022lst}, we select the best learning rate in the dev set between the values proposed in the original papers, $[5e^{-4}, 4e^{-4}, 3e^{-4}, 2e^{-4}]$ and $[3e^{-4}, 1e^{-3}, 3e^{-3}]$, respectively. We do not use the initialization setup proposed in LoRA or LST nor do we drop any layers for the LST method.}
\label{table:glue}

	\setlength{\tabcolsep}{1pt}    
    \begin{tabularx}{\textwidth}{lc|YYYYYYY|Y}
    \toprule
    Method &  Param. (\%) & CoLA & SST-2 & MRPC & QQP & QNLI & MNLI & STS-B & Avg. $\uparrow$\\
    \midrule
    Avg. \# Tokens & --- & 11.3 & 13.3 & 53.2 & 30.6 & 49.4 & 39.8 & 27.8 & 32.2 \\
\addlinespace
    Full Fine-Tuning$^{\dagger}$ & 100.0 & 60.7 & \textbf{94.6} & 88.3 & \textbf{72.0} & 92.4 & 85.8 & 85.8 & 82.8 \\
    Adapters & 3.6 & 59.5 & 94.0 & 89.5 & 71.8 & 90.7 & 84.9 & \textbf{86.9} & 82.5 \\
BitFit & 0.1 & 59.7 & 94.2 & 88.9 & 70.5 & 92.0 & 84.5 & 85.0 & 82.1  \\
    Diff Pruning & 0.5 & \textbf{61.1} & 94.1 & \textbf{89.7} & 71.1 & \textbf{93.3} & \textbf{86.4} & 86.0 & \textbf{83.1}  \\
    \addlinespace
    Ladder Side Tuning$^{\dagger}$ & 2.4 & 56.4 & 93.4 & 88.0 & 66.9 & 89.1 & 82.9 & 86.6 & 80.5  \\
    LoRA$^{\dagger}$ &  0.3 & 58.5 & 94.0 & 89.2 & 71.1 & 91.1 & 84.7 & 84.6 & 81.9  \\
\rowcolor{lightcyan}
    \method$^{\dagger}$ & 100.0 & 59.6 & 93.9 & 88.0 & 70.8 & 91.0 & 85.4 & 86.0 & 82.1 \\
    \bottomrule
    \end{tabularx}
\end{table*}

\subsection{\method for Attention Layers}
\label{method:att}

For attention layers, we compute the attention as:
\begin{gather}
\left[Q_{\mathcal{G}}, K_{\mathcal{G}}, V_{\mathcal{G}}\right] = h_{\mathcal{G}}W_{\left[Q,K,V\right]}  + b_{\left[Q,K,V\right]}
\label{transformer:v}\\
\left[Q_{\bar{\mathcal{G}}}, K_{\bar{\mathcal{G}}}, V_{\bar{\mathcal{G}}}\right] = h_{\bar{\mathcal{G}}} W_{\left[Q,K,V\right]}  + b_{\left[Q,K,V\right]}
\label{transformer:v-no-grad}\\
h_{\mathcal{G}} = \softmax \left( \nicefrac{Q_{\mathcal{G}} \left[K_{\bar{\mathcal{G}}}, K_{\mathcal{G}}\right]^{\top}}{\sqrt{d}}  \right) \left[V_{\bar{\mathcal{G}}}, V_{\mathcal{G}}\right] \label{transformer:att-grad}\\
h_{\bar{\mathcal{G}}} = \softmax \left( \nicefrac{Q_{\bar{\mathcal{G}}} \left[K_{\bar{\mathcal{G}}}, K_{\mathcal{G}}\right]^{\top}}{\sqrt{d}}  \right) \left[V_{\bar{\mathcal{G}}}, V_{\mathcal{G}}\right] \label{transformer:att-no-grad}
\end{gather}
where  $W_{\left[Q,K,V\right]} \in \mathbb{R}^{d\times 3d}$ denotes the concatenated weights for the queries, keys, and values.
For the computation of un-selected positions in Eq.~\ref{transformer:v-no-grad} and Eq.~\ref{transformer:att-no-grad}, we again disable the gradient computation in PyTorch.
Algorithm~\ref{alg:tokentune} illustrates the steps for the forward pass of a transformer model with the proposed \method algorithm described in \Cref{method:att,method:dense}.

\begin{algorithm}[t]
	\setstretch{1.1}  \setlength{\lineskip}{3pt}
\small
\caption{\method (We omit layer normalization, skip connections, non-linear functions, and multi-head attention for simplicity)}
	\label{alg:tokentune}
\SetAlgoVlined  \SetKwBlock{WithNoGrad}{with \text{\normalfont{torch.no\_grad():}}}{end}
	\KwIn{input sequence $X$
	}
	\KwOut{$h_{\mathcal{G}}, h_{\bar{\mathcal{G}}}$
	}
\BlankLine
	Compute input token embeddings $h$ \\
	Re-organize input tokens into two groups ($h_{\mathcal{G}}$ and $h_{\bar{\mathcal{G}}}$)
	\BlankLine
	
\For{{layer} $ \textbf{\textup{in}}~\textup{transformers' layers} $}{
		\tcp{Compute the attention layer}
$ \left[Q_{\mathcal{G}}, K_{\mathcal{G}}, V_{\mathcal{G}}\right] = h_{\mathcal{G}}W_{\left[Q,K,V\right]}  + b_{\left[Q,K,V\right]}$  \\
$ h_{\mathcal{G}}  = \softmax \left( \frac{Q_{\mathcal{G}} \left[K_{\bar{\mathcal{G}}}, K_{\mathcal{G}}\right]^{\top}}{\sqrt{d}}  \right) \left[V_{\bar{\mathcal{G}}}, V_{\mathcal{G}}\right] $
		
		\BlankLine
		\WithNoGrad{
$ \left[Q_{\bar{\mathcal{G}}}, K_{\bar{\mathcal{G}}}, V_{\bar{\mathcal{G}}}\right] = h_{\bar{\mathcal{G}}} W_{\left[Q,K,V\right]}  + b_{\left[Q,K,V\right]} $ \\
$ h_{\bar{\mathcal{G}}} = \softmax \left( \frac{Q_{\bar{\mathcal{G}}} \left[K_{\bar{\mathcal{G}}}, K_{\mathcal{G}}\right]^{\top}}{\sqrt{d}}  \right) \left[V_{\bar{\mathcal{G}}}, V_{\mathcal{G}}\right] $
		}
		
		\BlankLine
		\tcp{Compute the feed-forward layer}
            $h_{\mathcal{G}} = h_{\mathcal{G}}W_1 + b_1$ \\
            $h_{\mathcal{G}} = h_{\mathcal{G}}W_2 + b_2$ \\
            \WithNoGrad{
                $h_{\bar{\mathcal{G}}} = h_{\bar{\mathcal{G}}}W_1 + b_1$ \\
                $h_{\bar{\mathcal{G}}} = h_{\bar{\mathcal{G}}}W_2 + b_2$
            }
	}
	Re-organize input tokens into the original order
\end{algorithm}

\begin{figure*}[t]
     \centering
     \hspace*{\fill}\begin{subfigure}[b]{.44\linewidth}
         \centering
         \includegraphics[width=\linewidth]{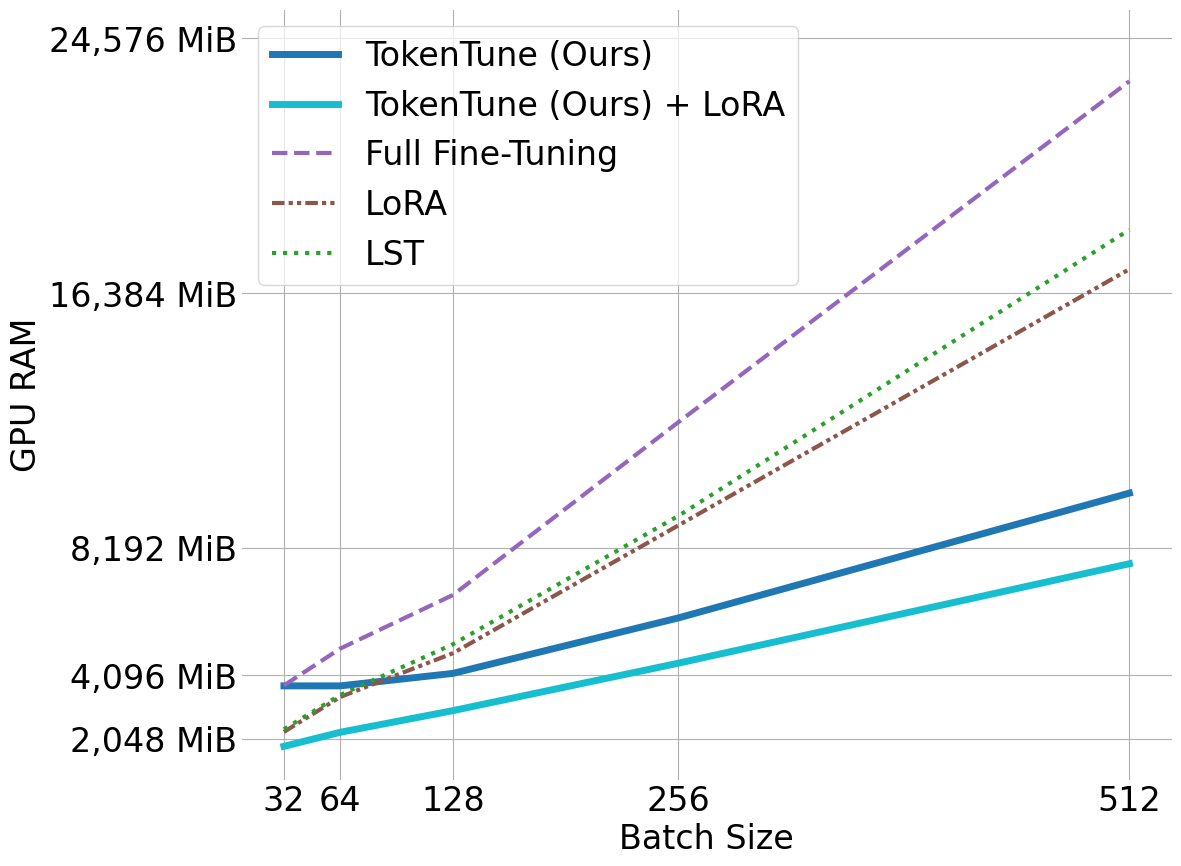}
\label{fig:ram}
     \end{subfigure}
     \hfill
     \begin{subfigure}[b]{.44\linewidth}
         \centering
         \includegraphics[width=\linewidth]{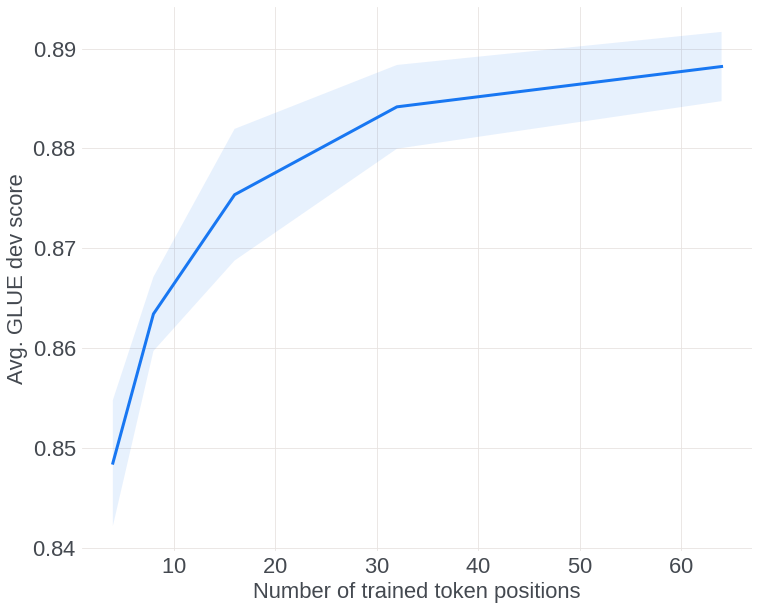}
\label{fig:prefix length}
     \end{subfigure}
     \hspace*{\fill}\caption{(left) We plot the GPU memory required to train \textsc{Bert}-base on the CoLA task given varying batch sizes. We compare our approach with two PEFT approaches: Ladder Side Tuning (LST) and LoRA. 
    (right) We plot the  mean and standard deviation performance on the dev set of five runs when training \textsc{Bert}-base on two tasks from the GLUE benchmark: MRPC and STS-B. We use our memory efficient fine-tuning approach with a different number of selected input tokens for the gradient computation.}
    \label{fig:graphs}
\end{figure*}

\section{Application to Medium-Size Encoders}  \label{sec:exp:medium}

Alternative methods such as zero-shot learning or prompting usually underperform fine-tuning \citep{DBLP:conf/nips/BrownMRSKDNSSAA20}. Thus, in many cases, fine-tuning medium size language models may offer a better balance in terms of cost and performance, compared with fine-tuning large language models (LLMs) or conditioning their outputs with prompt approaches \citep{DBLP:conf/iclr/LiTLH22, DBLP:conf/naacl/SchickS21}. Medium-size models may also be used as individual components, co-trained to encode information for a larger system \citep{pfeiffer_23}. Finally, as detailed in \Cref{sec:app:memorycomplexity:bkd}, the distribution of the GPU memory usage may be very different given the order of magnitude of the fine-tuned model's number of parameters. For large-size models, the majority of the memory is often dedicated to storing parameters and optimizer states, thus maximizing the relevance of PEFT approaches. For medium-size language models, fine-tuned with large batch sizes, the majority of the memory may be dedicated to storing the intermediate activation, thus maximizing the impact of \method.

\subsection{Downstream Task Performance}

We first validate the relevance of our method on the GLUE benchmark \citep{wang_18}. 
We use a similar hyper-parameter search space as in \citep{zaken_22}, by performing a cross validation on the dev set using a learning rate in $[5e^{-5}, 3e^{-5}, 2e^{-5}, 1e^{-5}]$. We set the batch size to $16$ and perform $3$ epochs on large datasets and $20$ epochs on small ones (MRPC, STS-B, CoLA). We use \textsc{Bert}-large \citep{devlin_19} and either fine-tune the model fully, or use \method and propagate the gradient through $16$ input positions. We then evaluate our model on the test set and report the results in Table~\ref{table:glue}.

As shown in the second part of Table~\ref{table:glue}, the average GLUE score of \method is comparable to that of full fine-tuning, thus empirically validating the effectiveness of our approach.Table~\ref{table:glue} also shows that \method either outperforms or performs similarly to existing SOTA approaches.
Precisely speaking, the performance of these memory-efficient fine-tuning methods, including \method, is often slightly worse than that of full fine-tuning.
In comparison to full fine-tuning, some amount of performance loss with these methods is expected as 
they approximate or simplify the optimization process of full fine-tuning to reduce memory footprint.
We hypothesize that some tasks, such as QQP and QNLI, are more difficult, or sensitive to overfitting than others,
given that updating a small proportion of model parameters or using only a subset of input tokens for gradient computation 
achieves suboptimal performances on those tasks in most cases.
The former case would require the development of sophisticated techniques to more effectively select a subset of parameters or input tokens to optimize,
while the latter case may benefit from the use of regularization techniques for neural networks, including \citet{DBLP:conf/iclr/GoukHP21,DBLP:conf/iclr/ForetKMN21,DBLP:conf/nips/LiZ21},
the investigation of which we leave for future studies.

\subsection{Ratio of Tuned Input Positions} \label{sec:medium:ratio}

Given our token-selective fine-tuning approach, we then evaluate the impact of the number of frozen input positions on the performance. We use our selective procedure to fine-tune \textsc{Bert}-base on two tasks from the GLUE benchmark: MRPC and STS-B. 
We set the hyper-parameters as follows: $5e^{-5}$ for the learning rate, $32$ for the batch size and $4$ epochs.
We use different values for $k$ (i.e., the number of trained input positions), ranging between $4$ and $64$. We report in Figure~\ref{fig:graphs} (right), the average performance on the dev set of the tasks.\footnote{We provide some descriptive statistics in Appendix~\ref{app:stats_desc} to better understand how the absolute number of frozen input positions relates with the relative number of frozen input positions. The statistics include distribution of the sentence length for the two subtasks (MRPC and STS-B) used to produce Figure~\ref{fig:graphs} (right).}

As seen in Figure~\ref{fig:graphs}, the performance increases from $84.8$ to $88.8$ as the number of trained positions increases from $4$ to $64$. However, by only tuning $32$ positions, we already reach an average performance of $88.4$, close to the $88.8$ obtained by training $64$ input positions.
Our method surpasses the performance of freezing some bottom layers, as shown in \citep{lee_19}, where only tuning the four bottom layers resulted in a 10\% decrease in performance on the GLUE benchmark.

\begin{table*}[t]
    \centering
\caption{Few-shot evaluation on question-answering benchmarks including: AI2 Reasoning Challenge (25-shot) \citep{clark_18}, MMLU (5-shot) \citep{hendrycks_21}, HellaSwag (10-shot) \citep{zellers19}, TruthfulQA (0-shot) \citep{lin_22}, and WinoGrande (0-shot) \citep{DBLP:conf/aaai/SakaguchiBBC20}. We use the evaluation scripts and prompt formatting from the "Language Model Evaluation Harness" \citep{eval-harness}. We report the average accuracy on five MMLU ethics tasks and WinoGrande, the normed accuracy on ARC and HellaSwag, and the MC2 score on TruthfulQA. 
   	We indicate in \textbf{bold} the best result for each task.
   	We report the results with the raw Llama2-7B model \citep{touvron_23} and the Llama2-7B fine-tuned on the Platypus curated instruction dataset \citep{lee_23} using LoRA \citep{hu_22}, QLoRA \citep{dettmers_23} and the proposed \method. When fine-tuning with \method, we select 30\% of the tokens for the gradient computation.}
    \label{tab:llm-perf}

	\setlength{\tabcolsep}{1pt}

\resizebox{.98\textwidth}{!}{

\begin{tabularx}{\textwidth}{lYYYYYY}
	\toprule
	\makecell[c]{{Method}} & {MMLU} & {ARC} & \makecell[c]{{Hella}\\{Swag}} & \makecell[c]{{Truthful}\\{QA}} & \makecell[c]{{Wino}\\{Grande}} & {Avg. $\uparrow$}\\
	\midrule
	Llama 7B & 64.44 & 52.39 & \textbf{78.97} & 38.97 & 68.90 & 60.73\\\midrule
	Llama 7B w/ LoRA & \textbf{65.89} & 55.38 & 78.76 & 42.64 & 68.35 & 62.20\\
	\rowcolor{lightcyan}
	Llama 7B w/ LoRA+\method (Ours) & 65.42 & 54.01 & 78.82 & \textbf{43.78} & 68.35 & 62.08\\\midrule
	Llama 7B w/ QLoRA & 65.08 & \textbf{56.06} & 78.60 & 43.64 & 69.38 & \textbf{62.55}\\
	\rowcolor{lightcyan}
	Llama 7B w/ QLoRA+\method (Ours) & 65.78 & 53.92 & 78.74 & 41.91 & 69.38 & 61.95\\\midrule
	\rowcolor{lightcyan}
	Llama 7B w/ \method (Ours) & 63.06 & 53.07 & 77.90 & 42.18 & \textbf{69.93} & 61.23\\
	\bottomrule
\end{tabularx}

} 

\end{table*}

\subsection{GPU Memory Impact} \label{sec:medium:mem}

Finally, we analyze the GPU memory required to fine-tune models using various approaches.  We train our \textsc{Bert}-base model for $100$ steps on the CoLA task using various batch sizes and report the peak GPU memory used. We compare with two other PEFT fine-tuning approaches close to ours: Ladder Side Tuning \citep{sung2022lst} and LoRA \citep{hu_22}. LoRA freezes most of the model parameters, while only training additional low-rank matrices, whose weights are added to the backbone network. Ladder Side Tuning (LST) freezes the model parameters but trains a side-network with smaller dimensions, taking as input intermediate activations from the backbone model.

\Cref{fig:graphs} shows the evolution of the required GPU memory with respect to the batch size. GPU memory increases with the batch size for every approach. \method is more memory efficient by a large margin. When using a batch size of $512$, it requires two times less memory than full fine-tuning: $23,196$ MiB needed for full fine-tuning is reduced to $9,952$ MiB with our method.

All methods minimize GPU memory usage. LoRA and LST reduce the memory required to store optimizer states and parameter gradients, while our method reduces  the memory for storing intermediate activations. 
Interestingly enough, it is possible to use these approaches in conjunction to reduce the memory for all three contributions. 
Fig.~\ref{fig:graphs} shows that we can further reduce the memory by combining \method with LoRA, thus requiring only $7,682$ MiB with a batch size of 512, a third of the memory used for full fine-tuning.

\vspace{-0.5em}
\section{Application to Large-Size Decoders}  \label{sec:exp:large}
\vspace{-0.5em}

We also seek to evaluate our method on larger size pre-trained language models (LLMs).

\subsection{Instruction Tuning and Few-Shot Evaluation} 

LLMs are typically further fine-tuned on curated datasets to tailor them to specific domains and enhance their capacity to follow instructions \citep{wang_23, alpaca, mukherjee_23}. In this section, we employ instruction tuning on these datasets to fine-tune the LLMs and then assess the performance of the resulting models using few-shot benchmarks.

\paragraph{Instruction Tuning.} 
We fine-tune the \mbox{Llama2-7B} model~\citep{touvron_23}
via instruction tuning with the Open-Platypus\footnote{\url{https://huggingface.co/datasets/garage-bAInd/Open-Platypus}}~\citep{lee_23} dataset.
Note that, while Open-Platypus consists of 11 open-source datasets, 
we exclude two of them\footnote{\texttt{leetcode-solutions-python-testgen-gpt4} and \texttt{airoboros-gpt4-1.4.1}} that include outputs from GPT~\cite{DBLP:journals/corr/abs-2303-08774}, and instead use the other nine datasets for fine-tuning.

\paragraph{Hyper-Parameter Settings.} We conduct all experiments in this section on Nvidia H100 GPU. Following \citet{lee_23}, we fine-tune the model for one epoch, and use a learning rate of $4e^{-4}$ for LoRA~\citep{hu_22} and QLoRA~\citep{dettmers_23}, and $4e^{-5}$ otherwise. We use a batch size of 1 with 32 gradient accumulation steps. We apply the adapters on the feed-forward modules from each layer, following the method described in \citet{he_22}. We prompt the model without step-wise reasoning using the Alpaca~\citep{alpaca} prompt template detailed in \Cref{app:instruct}.

\paragraph{Few-Shot Evaluation.} 
Then, we evaluate our method against other memory-efficient fine-tuning approaches by assessing its performance on several few-shot benchmarks, such as MMLU \citep{hendrycks_21}, ARC easy and challenge \citep{clark_18}, HellaSwag \citep{zellers19}, TruthfulQA \citep{lin_22}, and WinoGrande \citep{DBLP:conf/aaai/SakaguchiBBC20}. We utilize the evaluation scripts provided by the "Language Model Evaluation Harness" \citep{eval-harness}. During the evaluation process, the model outputs the probability associated with closed-form problems defined by the context, question, and multiple potential answers. We select the answer choice with the text associated with the highest probability. 

Table~\ref{tab:llm-perf} reports the accuracy of the model output against the ground truth answer. 
Our method achieves competitive performance gains that are comparable to the performance improvements obtained by other memory efficient fine-tuning approaches. 
We are able to improve the evaluation accuracy upon the base LLama2-7B model, increasing the average accuracy from 60.7 to 61.2. We observe the most significant improvements for TruthfulQA (+3.2) and WinoGrande (+1.0) tasks. We also combine \method with LoRA and QLoRA, further improving the evaluation accuracy compared to the use of \method alone.

\begin{table}[t]
 	\centering
\caption{
   Few-shot evaluation results and peak memory usage (GiB) as Llama2-7B is fine-tuned on instruction datasets 
   with (a) \method, (b) \method + LoRA and (c) \method + QLoRA, varying the selection ratio of input tokens. Best results in \textbf{bold}.
 }
 	\label{tab:ablation:tokenratio}
 	\setlength{\tabcolsep}{0.3em}

        \begin{subtable}[c]{0.5\textwidth}
 		\centering
 		\caption{\method{}}
 		\label{tab:ablation:tokenratio:sft}
 		\resizebox{.95\textwidth}{!}{\begin{tabular}{c|c|cccccc}
 			\toprule
 			\makecell[c]{\textbf{Selection}\\\textbf{Ratio}} & \makecell[c]{\textbf{Peak}\\\textbf{Mem.}} & \textbf{MMLU} & \textbf{ARC} & \makecell[c]{\textbf{Hella}\\\textbf{Swag}} & \makecell[c]{\textbf{Truthful}\\\textbf{QA}} & \makecell[c]{\textbf{Wino}\\\textbf{Grande}} & \makecell[c]{\textbf{Avg.}\\\textbf{Perf.}}\\
 			\midrule
 			10\% & \textbf{64.40} & 61.56 & 51.71 & 78.35 & 41.88 & 70.01 & 60.70\\
 			20\% & 65.08 & \textbf{65.01} & 52.65 & \textbf{78.37} & 42.02 & 69.46 & \textbf{61.50}\\
 			30\% & 65.94 & 63.06 & \textbf{53.07} & 77.90 & \textbf{42.18} & 69.93 & 61.23\\
 			40\% & 68.42 & 63.78 & 52.90 & 77.90 & 41.45 & \textbf{70.32} & 61.27\\
 			50\% & 74.32 & 62.98 & 52.73 & 78.32 & 42.11 & 69.38 & 61.10\\
 			\bottomrule
 		\end{tabular}
 		}\end{subtable}
 	
 	\vspace{0.5em}
	
 	\begin{subtable}[c]{0.5\textwidth}
 		\centering
 		\caption{\method{} + LoRA}
 		\label{tab:ablation:tokenratio:lora+sft}
 		\resizebox{.95\textwidth}{!}{\begin{tabular}{c|c|cccccc}
 			\toprule
 			\makecell[c]{\textbf{Selection}\\\textbf{Ratio}} & \makecell[c]{\textbf{Peak}\\\textbf{Mem.}} & \textbf{MMLU} & \textbf{ARC} & \makecell[c]{\textbf{Hella}\\\textbf{Swag}} & \makecell[c]{\textbf{Truthful}\\\textbf{QA}} & \makecell[c]{\textbf{Wino}\\\textbf{Grande}} & \makecell[c]{\textbf{Avg.}\\\textbf{Perf.}}\\
 			\midrule
 			10\% & \textbf{45.47} & 64.17 & \textbf{54.44} & 78.68 & 38.77 & \textbf{69.61} & 61.13\\
 			20\% & 48.21 & 65.41 & 54.35 & \textbf{79.01} & 42.21 & 69.38 & 62.07\\
 			30\% & 52.77 & 65.42 & 54.01 & 78.82 & \textbf{43.78} & 68.35 & 62.08\\
 			40\% & 56.31 & 64.35 & 52.65 & 78.69 & 41.05 & 68.90 & 61.13\\
 			50\% & 64.34 & \textbf{65.87} & 54.01 & 78.68 & 42.46 & 69.38 & \textbf{62.08}\\
 			\bottomrule
 		\end{tabular}
 		}\end{subtable}
 	
 	\vspace{0.5em}
 	
 	\begin{subtable}[c]{0.5\textwidth}
 		\centering
 		\caption{\method{} + QLoRA}
 		\label{tab:ablation:tokenratio:qlora+sft}
 		\resizebox{.95\textwidth}{!}{\begin{tabular}{c|c|cccccc}
			\toprule
			\makecell[c]{\textbf{Selection}\\\textbf{Ratio}} & \makecell[c]{\textbf{Peak}\\\textbf{Mem.}} & \textbf{MMLU} & \textbf{ARC} & \makecell[c]{\textbf{Hella}\\\textbf{Swag}} & \makecell[c]{\textbf{Truthful}\\\textbf{QA}} & \makecell[c]{\textbf{Wino}\\\textbf{Grande}} & \makecell[c]{\textbf{Avg.}\\\textbf{Perf.}}\\
			\midrule
			10\% & \textbf{11.47} & 63.54 & 54.18 & 78.58 & 39.79 & 68.98 & 61.02\\
			20\% & 15.68 & 64.05 & 53.92 & \textbf{78.81} & 40.33 & \textbf{69.85} & 61.39\\
			30\% & 19.71 & \textbf{65.78} & 53.92 & 78.74 & 41.91 & 69.38 & \textbf{61.95}\\
			40\% & 24.11 & 64.85 & \textbf{54.35} & 78.70 & \textbf{41.98} & 69.14 & 61.80\\
			50\% & 31.06 & 65.29 & 53.75 & 78.70 & 40.63 & 69.06 & 61.49\\
			\bottomrule
		\end{tabular}
 		}\end{subtable}
\end{table}

\subsection{Ratio of Tuned Input Positions}

As done for medium-size encoders in \Cref{sec:medium:ratio}, we then evaluate the impact of the ratio of tuned input positions on the few-shot accuracy. 
We measure the few-shot accuracy of Llama2-7B models fine-tuned using \method with varying ratio of tuned input positions. 
\Cref{tab:ablation:tokenratio} shows few-shot evaluation accuracy of Llama2-7B when the ratio of fine-tuned positions ranges from 10\% to 50\% .

Contrary to what we observed in \Cref{sec:medium:ratio}, we do not necessarily observe a strong correlation between the few-shot accuracy and the ratio of tuned positions. 
In fact, we obtain the best performances most often when 20\%--30\% of input positions are fine-tuned. 
It is important to observe that the average sequence length in these experiments far exceeds the one from the experiments on the GLUE benchmark. This suggests that tuning a relatively small number of positions may be sufficient to successfully fine-tune the model on specific datasets. 

\begin{figure}[t]
\centering
     \includegraphics[width=1.0\linewidth]{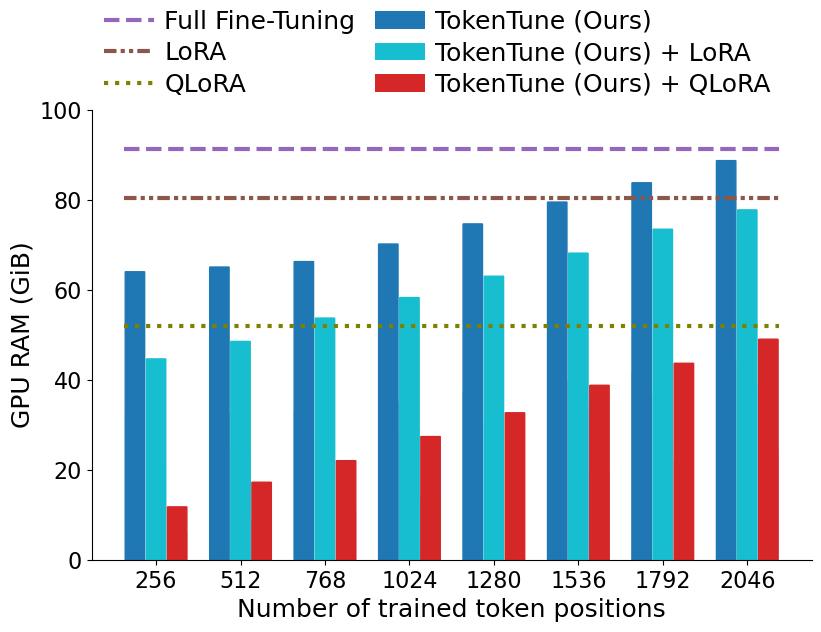}
\caption{GPU memory required to fine-tune Llama2-7B~\citep{touvron_23}. We measure the memory by fine-tuning the model on artificially generated data with a given sequence length and batch size. We set the batch size to 1 and the sequence length to \numprint{2048}. We show the memory usage when combining  \method with LoRA and QLoRA and plot the evolution of the memory required to fine-tune the model on a H100 GPU with a number of trained positions ranging between \numprint{256} and \numprint{2046} (we leave at least 2 positions not tuned). 
    Since we could not perform full fine-tuning on our hardware, we estimate the full fine-tuning memory based on the memory reported for \method and LoRA.
    Specific memory usage values can be found in \Cref{tab:gpu_mem_usage}.}
    \label{fig:llm-memory}
\end{figure}

\subsection{GPU Memory Impact}
\label{sec:exp:large:gpu_memory}

As in \Cref{sec:medium:mem}, we analyze the impact of our method on the GPU memory required to fine-tune large language models. \Cref{fig:llm-memory} and \Cref{tab:ablation:tokenratio} report the GPU memory usage for fine-tuning Llama2-7B as the number of trained input tokens changes.
Given an input sequence of length \numprint{2048}, \Cref{fig:llm-memory} shows that our model reduces the memory usage by up to 28\%, from 89 GiB to 64 GiB when reducing the number of trained positions from \numprint{2046} to \numprint{256}. 

The advantage of the proposed method is that it can be combined with other memory saving methods. 
We measure the peak memory required to fine-tune LLama2-7B when combining \method with LoRA or QLoRA. 
Since these approaches target different parts of the memory footprint,
we observe cumulative savings when they are used together. 
When combining LoRA with \method, the peak memory ranges between 78 GiB to 45 GiB depending on the number of tuned positions. 
Similarly, when combining QLoRA with \method, the peak memory decreases from 49 GiB to 12 GiB as a smaller selection ratio is used.

Overall, \Cref{fig:llm-memory} and \Cref{tab:ablation:tokenratio} show that 
the performance of TokenTune is not very sensitive to the choice of token selection ratio, 
while the memory cost is significantly reduced with a smaller token selection ratio.
Based on these results, 
our recommendation is to use 20\%--30\% as the default token selection ratio, and 
test if further improvements in performance and memory usage can be obtained for the given task, with a smaller selection ratio.

\section{Conclusion}
\label{sec:conclusion}
In this paper, we propose \method, a method for reducing the GPU memory required to fine-tune transformer-based models, such as large language models. 
Our contributions are as follows.
\begin{itemize}[leftmargin=1em,topsep=-2pt,itemsep=-3pt]
	\item \textbf{Novelty.} \method is the first approach that reduces the GPU memory footprint for fine-tuning via token selection, 
	which selects a subset of the input positions through which the gradient is propagated, while keeping the others frozen.
	\item \textbf{Combinability.} The proposed token selection strategy can be combined with other memory- and parameter-efficient fine-tuning approaches, achieving a greater memory reduction together.
	\item \textbf{Effectiveness.} We empirically benchmark \method using large language models with up to billions of parameters. 
	As \Cref{fig:crownjewel} and \Cref{table:glue} show, \method achieves similar prediction accuracy to representative memory- and parameter-efficient methods, such as LoRA and QLoRA, 
	while significantly reducing the memory usage for fine-tuning (e.g., a joint application of \method and QLoRA uses 79\% less memory than full fine-tuning).
\end{itemize}

\clearpage
\section{Limitations}
\label{sec:limitations}

While \method effectively reduces the memory required for storing intermediate activations,
it does not affect the other parts of GPU memory usage, such as the one for parameter gradients.
However, as we showed in experiments, \method can be combined with memory-efficient methods that reduce those other parts of memory~footprint.
Also, the evaluation of \method in this work focused on one domain, namely, language models.
Given the applicability of \method to other domains, such as vision~\citep{DBLP:conf/iclr/DosovitskiyB0WZ21},
we hope to investigate its effectiveness in broader settings in the future.

\paragraph{Potential Risks.}
Since this paper presents a method for memory-efficient fine-tuning of transformer-based models, such as LLMs, and is not tied to particular applications,
we do not see potential risks of the proposed method.

{

}
\clearpage

\appendix

\section{Instruction Template}\label{app:instruct}
Regarding the instruction tuning of large LLMs, we prompt the model without step-wise reasoning using the Alpaca~\citep{alpaca} prompt template presented below.

\vspace{0.5em}
\noindent
\fbox{\ttfamily \begin{minipage}{1.0\columnwidth}
{``Below is an instruction that describes a task, paired with an input that provides further context. Write a response that appropriately completes the request.

\#\#\# Instruction:
\{instruction\}

\#\#\# Input:
\{input\}

\#\#\# Response:\\
''}
\end{minipage}}
\section{Software}\label{sec:app:software}
Here we provide the details of the software used for the implementation of \method 
as well as the fine-tuning and evaluation of \method and baselines.
Our implementation of \method builds upon the HuggingFace Transformers library\footnote{\url{https://github.com/huggingface/transformers}} (v4.33.1).
For LoRA~\citep{hu_22}, we used the HuggingFace PEFT library\footnote{\url{https://github.com/huggingface/peft}} (v.0.5.0).
Datasets used for fine-tuning were obtained from the HuggingFace Datasets library\footnote{\url{https://github.com/huggingface/datasets}} (v2.18.0).
We used Open-Platypus\footnote{\url{https://huggingface.co/datasets/garage-bAInd/Open-Platypus}} for fine-tuning.
For the evaluation with the Llama2 model in \Cref{sec:exp:large},
we used the lm-evaluation-harness framework\footnote{\url{https://github.com/EleutherAI/lm-evaluation-harness}} (v.0.4.2).
We used the PyTorch framework\footnote{\url{https://github.com/pytorch/pytorch}} (v.2.0.1). Results from \Cref{table:glue} are scored by the evaluation server.\footnote{\url{https://gluebenchmark.com/leaderboard}} As in \citet{devlin_19}, we discard results for the WNLI~task.\footnote{See (12) from \url{https://gluebenchmark.com/faq}}

\section{License}\label{sec:app:license}
The majority of \method is licensed under CC-BY-NC, however portions of the project are available under separate license terms: Transformers is licensed under the Apache 2.0 license.
The license of other libraries used for this paper is as follows.
The PEFT and Datasets libraries from HuggingFace are under the Apache-2.0 license.
The lm-evaluation-harness framework is under the MIT license.
PyTorch is under the modified BSD-3 license.
Open-Platypus used for fine-tuning consists of multiple datasets; their license information can be found at \url{https://huggingface.co/datasets/garage-bAInd/Open-Platypus}.

\section{Training and Evaluation Data} \label{data}

\textsc{Bert} model has been pre-trained on \numprint{3300}M words.
Regarding the instruction tuning experiments, we tuned the \mbox{Llama2-7B} on \numprint{21221} samples from the Open-Platypus~\citep{lee_23} dataset. Note that, while Open-Platypus consists of 11 open-source datasets, 
we exclude two of them\footnote{\texttt{leetcode-solutions-python-testgen-gpt4} and \texttt{airoboros-gpt4-1.4.1}} that include outputs from GPT~\cite{DBLP:journals/corr/abs-2303-08774}, and instead use the other nine datasets for fine-tuning. Llama2-7B has been pre-trained on 2T tokens and fine-tuned on \numprint{100000} samples.\footnote{\url{https://llama.meta.com/llama2/}}

\section{Memory Breakdown}\label{sec:app:memorycomplexity:bkd}

Parameter-Efficient Fine-Tuning (PEFT) approaches aim at reducing the compute and storage requirements to fine-tune LLMs by only updating a small subset of the model parameters. 
As a result, we do not need to store any corresponding gradients and optimizer states for the frozen parameters. 
When parameters, gradients, and optimizer states represent the majority of the GPU memory usage,
these PEFT methods can effectively reduce the memory cost.
However, when most GPU memory is used to store intermediate activations, 
which are required for gradient computation during the backward pass, 
these PEFT methods cannot effectively cut down the memory cost.

\begin{table*}[t!]
    \centering
    \small
\caption{GPU memory required to fine-tune Llama2-7B~\citep{touvron_23} using \method with a varying selection ratio, as well as QLoRA and LoRA. 
    	Since we could not perform full fine-tuning on our hardware, we estimate the full fine-tuning memory based on the memory reported for \method,  \method + LoRA, and LoRA.
    	See \Cref{sec:exp:large:gpu_memory} and \Cref{fig:llm-memory} for details of the experiment.}
    \label{tab:gpu_mem_usage}

	\setlength{\tabcolsep}{1pt}    
    \begin{tabularx}{1.0\textwidth}{YYYYYYY}
    \toprule
    Selection Ratio & \method (Ours) + QLoRA & QLoRA & \method (Ours) + LoRA & LoRA & \method (Ours) & Full Fine-Tuning \\
    \midrule
    12.5\% & 11.7 GiB & 51.9 GiB & 44.6 GiB & 80.4 GiB & 64.0 GiB & 91.4 GiB \\
    25.0\% & 17.2 GiB & 51.9 GiB & 48.5 GiB & 80.4 GiB & 65.0 GiB & 91.4 GiB \\
    37.5\% & 22.0 GiB & 51.9 GiB & 53.7 GiB & 80.4 GiB & 66.3 GiB & 91.4 GiB \\
    50.0\% & 27.4 GiB & 51.9 GiB & 58.3 GiB & 80.4 GiB & 70.2 GiB & 91.4 GiB \\
    62.5\% & 32.7 GiB & 51.9 GiB & 63.0 GiB & 80.4 GiB & 74.6 GiB & 91.4 GiB \\
    75.0\% & 38.8 GiB & 51.9 GiB & 68.1 GiB & 80.4 GiB & 79.5 GiB & 91.4 GiB \\
    87.5\% & 43.7 GiB & 51.9 GiB & 73.4 GiB & 80.4 GiB & 83.8 GiB & 91.4 GiB \\
    99.9\% & 49.0 GiB & 51.9 GiB & 77.7 GiB & 80.4 GiB & 88.7 GiB & 91.4 GiB \\
\bottomrule
    \end{tabularx}
\end{table*}

Table~\ref{table: Distribution of the GPU memory.} presents the GPU memory required to perform one training step with \textsc{Bert}-base \citep{devlin_19} and OPT \citep{zhang_22} on a consumer hardware GPU. We calibrate the example such that the memory requirement is roughly the same for both models. In this configuration we can only fit a single example for OPT, while we can use a batch size of \numprint{256} for \textsc{Bert}. We observe that the memory breakdown is very different between the two configurations. The required memory drastically increases during the forward pass for \textsc{Bert} and during the backward pass for OPT. When comparing the execution of forward pass with and without enabling gradient computation in PyTorch, we estimate that the memory cost to store intermediate activations represents around 22 Gb for \textsc{Bert} and less than 1 Gb for OPT. On the contrary, we estimate that computing and storing the parameter gradients increase the memory requirement by less than 1 Gb for \textsc{Bert} and around 5 Gb for OPT. When applying LoRA \citep{hu_22}, a PEFT method, we observe that the memory drastically decreases for OPT, while having a less significant impact on \textsc{Bert}.
These examples demonstrate that an effective memory reduction across different usage scenarios
can be achieved by combining a suite of memory-efficient fine-tuning methods that can complement each other by reducing different parts of the memory footprint simultaneously.

\begin{table}[t!]
	\caption{Using two models requiring roughly the same GPU memory, we observe that the memory breakdown and the impact of PEFT methods application are very different. For each model, we show the evolution of the GPU memory ($\times10^3$ MiB) required for performing one training step for OPT-1B3 \citep{zhang_22} with a batch size of 1 and a sequence length of \numprint{128} and \textsc{Bert}-base \citep{devlin_19} with a batch size of \numprint{256}, a sequence length of \numprint{128}. Fwd (w/o grad) corresponds to the execution of the forward pass, while disabling gradient computation.}
	\label{table: Distribution of the GPU memory.}
\setlength{\tabcolsep}{2pt}
    \begin{tabularx}{0.48\textwidth}{lYY|YY} \toprule
         & \multicolumn{2}{c}{}  & \multicolumn{2}{c}{w/ LoRA}\\
         & \textsc{Bert} & OPT  & \textsc{Bert} & OPT \\
        \midrule
Cuda Context & \numprint{0.8} & \numprint{0.8}  & \numprint{0.8} & \numprint{0.8}\\
+ Model weights & \numprint{1.3} &\textbf{\numprint{5.8}} & \numprint{1.3} &\textbf{\numprint{5.8}}\\
+ Fwd (w/o grad) & \numprint{2.9} &\numprint{6.1}&  \numprint{2.9} &\numprint{6.1}\\
+ Fwd (w/ grad) & \textbf{\numprint{24.8}} & \numprint{6.3}&\textbf{\numprint{20.6}}&\numprint{6.3} \\
+ Bwd & \numprint{25.2} & \textbf{\numprint{11.3}}&\numprint{21.0}& \numprint{6.3}\\
+ Optimizer step & \numprint{25.2} & \textbf{\numprint{21.4}}&\numprint{21.0}& \numprint{6.3}\\
\bottomrule
    \end{tabularx}
    \vspace{-0.75em}
\end{table}

\section{MRPC and STS-B Descriptive Statistics}\label{app:stats_desc}

\begin{table}[t!]
    \centering
    \small
    \caption{Distribution of the sentence length for the two GLUE subtasks (MRPC and STS-B).}
    \label{table:table_stats}
    \setlength{\tabcolsep}{1pt}
    \begin{tabularx}{\linewidth}{YYYYYY}
    \toprule
    Task & 25th percentile (P25\%) & Avg. tokens per sentence & 75th percentile (P75\%) & Max tokens per sentence & \# Training Sentences \\\midrule
    STS-B & 19.0 &  27.8 & 31.0 & 125 & 5,749 \\
    MRPC & 44.0 & 53.2 & 62.0 & 103 & 3,668 \\
    \bottomrule
    \end{tabularx}
\end{table}

\begin{table}[t!]
    \centering
    \small
    \caption{Relative proportion of fine-tuned tokens averaged over MRPC and STS-B tasks with respect to the number of fine-tuned tokens, along with the corresponding average performance (reported in Figure~\ref{fig:graphs}  (right)).}
    \label{table:table_stats_2}
    \setlength{\tabcolsep}{1pt}
    \begin{tabularx}{\linewidth}{YYY}
    \toprule
    \# Fine-Tuned Tokens & Average Relative Proportion of Fine-Tuned Tokens & Average Perf. \\\midrule
    4 &  13.6\% & 84.9 \\
    8 &  27.2\% & 86.4 \\
    16&  53.9\% & 87.6 \\
    32&  81.4\% & 88.4 \\
    64&  99.0\% & 88.8 \\
    \bottomrule
    \end{tabularx}
\end{table}

\Cref{table:table_stats} describes the relation between the absolute and relative number of frozen input positions. The statistics include distribution of the sentence length for the two subtasks (MRPC and STS-B) used to produce \Cref{fig:graphs} (right). We also report in \Cref{table:table_stats_2} the relative proportion of fine-tuned tokens averaged over MRPC and STS-B tasks, as the absolute number of fine-tuned tokens changes, along with the corresponding average performance, which is reported in \Cref{fig:graphs}  (right).

\vspace{0.45em}
\section{GPU Memory Usage}\label{app:gpu_mem_usage}
\vspace{0.05em}

\Cref{tab:gpu_mem_usage} shows the GPU memory usage required to fine-tune Llama2-7B~\citep{touvron_23} using the proposed \method with a varying selection ratio, as well as QLoRA and LoRA.
\Cref{fig:llm-memory} also visualizes the same results. See \Cref{sec:exp:large:gpu_memory} and \Cref{fig:llm-memory} for further details of the experiment.


\begin{thebibliography}{73}
\providecommand{\natexlab}[1]{#1}

\bibitem[{Aghajanyan et~al.(2021)Aghajanyan, Gupta, and
  Zettlemoyer}]{DBLP:conf/acl/AghajanyanGZ20}
Armen Aghajanyan, Sonal Gupta, and Luke Zettlemoyer. 2021.
\newblock \href {https://doi.org/10.18653/V1/2021.ACL-LONG.568} {Intrinsic
  dimensionality explains the effectiveness of language model fine-tuning}.
\newblock In \emph{Proceedings of the 59th Annual Meeting of the Association
  for Computational Linguistics and the 11th International Joint Conference on
  Natural Language Processing, {ACL/IJCNLP} 2021, (Volume 1: Long Papers),
  Virtual Event, August 1-6, 2021}, pages 7319--7328. Association for
  Computational Linguistics.

\bibitem[{Bai et~al.(2022)Bai, Jones, Ndousse, Askell, Chen, DasSarma, Drain,
  Fort, Ganguli, Henighan, Joseph, Kadavath, Kernion, Conerly, Showk, Elhage,
  Hatfield{-}Dodds, Hernandez, Hume, Johnston, Kravec, Lovitt, Nanda, Olsson,
  Amodei, Brown, Clark, McCandlish, Olah, Mann, and Kaplan}]{bai_22}
Yuntao Bai, Andy Jones, Kamal Ndousse, Amanda Askell, Anna Chen, Nova DasSarma,
  Dawn Drain, Stanislav Fort, Deep Ganguli, Tom Henighan, Nicholas Joseph,
  Saurav Kadavath, Jackson Kernion, Tom Conerly, Sheer~El Showk, Nelson Elhage,
  Zac Hatfield{-}Dodds, Danny Hernandez, Tristan Hume, Scott Johnston, Shauna
  Kravec, Liane Lovitt, Neel Nanda, Catherine Olsson, Dario Amodei, Tom~B.
  Brown, Jack Clark, Sam McCandlish, Chris Olah, Benjamin Mann, and Jared
  Kaplan. 2022.
\newblock \href {https://doi.org/10.48550/arXiv.2204.05862} {Training a helpful
  and harmless assistant with reinforcement learning from human feedback}.
\newblock \emph{CoRR}, abs/2204.05862.

\bibitem[{Brown et~al.(2020)Brown, Mann, Ryder, Subbiah, Kaplan, Dhariwal,
  Neelakantan, Shyam, Sastry, Askell, Agarwal, Herbert{-}Voss, Krueger,
  Henighan, Child, Ramesh, Ziegler, Wu, Winter, Hesse, Chen, Sigler, Litwin,
  Gray, Chess, Clark, Berner, McCandlish, Radford, Sutskever, and
  Amodei}]{DBLP:conf/nips/BrownMRSKDNSSAA20}
Tom~B. Brown, Benjamin Mann, Nick Ryder, Melanie Subbiah, Jared Kaplan,
  Prafulla Dhariwal, Arvind Neelakantan, Pranav Shyam, Girish Sastry, Amanda
  Askell, Sandhini Agarwal, Ariel Herbert{-}Voss, Gretchen Krueger, Tom
  Henighan, Rewon Child, Aditya Ramesh, Daniel~M. Ziegler, Jeffrey Wu, Clemens
  Winter, Christopher Hesse, Mark Chen, Eric Sigler, Mateusz Litwin, Scott
  Gray, Benjamin Chess, Jack Clark, Christopher Berner, Sam McCandlish, Alec
  Radford, Ilya Sutskever, and Dario Amodei. 2020.
\newblock \href
  {https://proceedings.neurips.cc/paper/2020/hash/1457c0d6bfcb4967418bfb8ac142f64a-Abstract.html}
  {Language models are few-shot learners}.
\newblock In \emph{Advances in Neural Information Processing Systems 33: Annual
  Conference on Neural Information Processing Systems 2020, NeurIPS 2020,
  December 6-12, 2020, virtual}.

\bibitem[{Chen et~al.(2016)Chen, Xu, Zhang, and
  Guestrin}]{DBLP:journals/corr/ChenXZG16}
Tianqi Chen, Bing Xu, Chiyuan Zhang, and Carlos Guestrin. 2016.
\newblock \href {https://arxiv.org/abs/1604.06174} {Training deep nets with
  sublinear memory cost}.
\newblock \emph{CoRR}, abs/1604.06174.

\bibitem[{Clark et~al.(2018)Clark, Cowhey, Etzioni, Khot, Sabharwal, Schoenick,
  and Tafjord}]{clark_18}
Peter Clark, Isaac Cowhey, Oren Etzioni, Tushar Khot, Ashish Sabharwal, Carissa
  Schoenick, and Oyvind Tafjord. 2018.
\newblock \href {https://arxiv.org/abs/1803.05457} {Think you have solved
  question answering? try arc, the {AI2} reasoning challenge}.
\newblock \emph{CoRR}, abs/1803.05457.

\bibitem[{Dai et~al.(2022)Dai, Dong, Hao, Sui, Chang, and
  Wei}]{DBLP:conf/acl/DaiDHSCW22}
Damai Dai, Li~Dong, Yaru Hao, Zhifang Sui, Baobao Chang, and Furu Wei. 2022.
\newblock \href {https://doi.org/10.18653/V1/2022.ACL-LONG.581} {Knowledge
  neurons in pretrained transformers}.
\newblock In \emph{Proceedings of the 60th Annual Meeting of the Association
  for Computational Linguistics (Volume 1: Long Papers), {ACL} 2022, Dublin,
  Ireland, May 22-27, 2022}, pages 8493--8502. Association for Computational
  Linguistics.

\bibitem[{Das et~al.(2023)Das, Zhang, Shi, Yin, and
  Zhang}]{DBLP:conf/emnlp/DasZS0Z23}
Sarkar Snigdha~Sarathi Das, Haoran Zhang, Peng Shi, Wenpeng Yin, and Rui Zhang.
  2023.
\newblock \href {https://doi.org/10.18653/V1/2023.EMNLP-MAIN.433} {Unified
  low-resource sequence labeling by sample-aware dynamic sparse finetuning}.
\newblock In \emph{Proceedings of the 2023 Conference on Empirical Methods in
  Natural Language Processing, {EMNLP} 2023, Singapore, December 6-10, 2023},
  pages 6998--7010. Association for Computational Linguistics.

\bibitem[{Dettmers et~al.(2022)Dettmers, Lewis, Belkada, and
  Zettlemoyer}]{dettmers_22}
Tim Dettmers, Mike Lewis, Younes Belkada, and Luke Zettlemoyer. 2022.
\newblock \href
  {http://papers.nips.cc/paper\_files/paper/2022/hash/c3ba4962c05c49636d4c6206a97e9c8a-Abstract-Conference.html}
  {{GPT3}.int8(): 8-bit matrix multiplication for transformers at scale}.
\newblock In \emph{Advances in Neural Information Processing Systems 35: Annual
  Conference on Neural Information Processing Systems 2022, NeurIPS 2022, New
  Orleans, LA, USA, November 28 - December 9, 2022}.

\bibitem[{Dettmers et~al.(2023)Dettmers, Pagnoni, Holtzman, and
  Zettlemoyer}]{dettmers_23}
Tim Dettmers, Artidoro Pagnoni, Ari Holtzman, and Luke Zettlemoyer. 2023.
\newblock \href
  {http://papers.nips.cc/paper\_files/paper/2023/hash/1feb87871436031bdc0f2beaa62a049b-Abstract-Conference.html}
  {{QLoRA}: Efficient finetuning of quantized llms}.
\newblock In \emph{Advances in Neural Information Processing Systems 36: Annual
  Conference on Neural Information Processing Systems 2023, NeurIPS 2023, New
  Orleans, LA, USA, December 10 - 16, 2023}.

\bibitem[{Devlin et~al.(2019)Devlin, Chang, Lee, and Toutanova}]{devlin_19}
Jacob Devlin, Ming{-}Wei Chang, Kenton Lee, and Kristina Toutanova. 2019.
\newblock \href {https://doi.org/10.18653/v1/n19-1423} {{BERT:} pre-training of
  deep bidirectional transformers for language understanding}.
\newblock In \emph{Proceedings of the 2019 Conference of the North American
  Chapter of the Association for Computational Linguistics: Human Language
  Technologies, {NAACL-HLT} 2019, Minneapolis, MN, USA, June 2-7, 2019, Volume
  1 (Long and Short Papers)}, pages 4171--4186. Association for Computational
  Linguistics.

\bibitem[{Dosovitskiy et~al.(2021)Dosovitskiy, Beyer, Kolesnikov, Weissenborn,
  Zhai, Unterthiner, Dehghani, Minderer, Heigold, Gelly, Uszkoreit, and
  Houlsby}]{DBLP:conf/iclr/DosovitskiyB0WZ21}
Alexey Dosovitskiy, Lucas Beyer, Alexander Kolesnikov, Dirk Weissenborn,
  Xiaohua Zhai, Thomas Unterthiner, Mostafa Dehghani, Matthias Minderer, Georg
  Heigold, Sylvain Gelly, Jakob Uszkoreit, and Neil Houlsby. 2021.
\newblock \href {https://openreview.net/forum?id=YicbFdNTTy} {An image is worth
  16x16 words: Transformers for image recognition at scale}.
\newblock In \emph{9th International Conference on Learning Representations,
  {ICLR} 2021, Virtual Event, Austria, May 3-7, 2021}. OpenReview.net.

\bibitem[{Foret et~al.(2021)Foret, Kleiner, Mobahi, and
  Neyshabur}]{DBLP:conf/iclr/ForetKMN21}
Pierre Foret, Ariel Kleiner, Hossein Mobahi, and Behnam Neyshabur. 2021.
\newblock \href {https://openreview.net/forum?id=6Tm1mposlrM} {Sharpness-aware
  minimization for efficiently improving generalization}.
\newblock In \emph{9th International Conference on Learning Representations,
  {ICLR} 2021, Virtual Event, Austria, May 3-7, 2021}. OpenReview.net.

\bibitem[{Gao et~al.(2021)Gao, Tow, Biderman, Black, DiPofi, Foster, Golding,
  Hsu, McDonell, Muennighoff, Phang, Reynolds, Tang, Thite, Wang, Wang, and
  Zou}]{eval-harness}
Leo Gao, Jonathan Tow, Stella Biderman, Sid Black, Anthony DiPofi, Charles
  Foster, Laurence Golding, Jeffrey Hsu, Kyle McDonell, Niklas Muennighoff,
  Jason Phang, Laria Reynolds, Eric Tang, Anish Thite, Ben Wang, Kevin Wang,
  and Andy Zou. 2021.
\newblock \href {https://doi.org/10.5281/zenodo.5371628} {A framework for
  few-shot language model evaluation}.

\bibitem[{Gheini et~al.(2021)Gheini, Ren, and May}]{DBLP:conf/emnlp/Gheini0M21}
Mozhdeh Gheini, Xiang Ren, and Jonathan May. 2021.
\newblock \href {https://doi.org/10.18653/V1/2021.EMNLP-MAIN.132}
  {Cross-attention is all you need: Adapting pretrained transformers for
  machine translation}.
\newblock In \emph{Proceedings of the 2021 Conference on Empirical Methods in
  Natural Language Processing, {EMNLP} 2021, Virtual Event / Punta Cana,
  Dominican Republic, 7-11 November, 2021}, pages 1754--1765. Association for
  Computational Linguistics.

\bibitem[{Gouk et~al.(2021)Gouk, Hospedales, and
  Pontil}]{DBLP:conf/iclr/GoukHP21}
Henry Gouk, Timothy~M. Hospedales, and Massimiliano Pontil. 2021.
\newblock \href {https://openreview.net/forum?id=IFqrg1p5Bc} {Distance-based
  regularisation of deep networks for fine-tuning}.
\newblock In \emph{9th International Conference on Learning Representations,
  {ICLR} 2021, Virtual Event, Austria, May 3-7, 2021}. OpenReview.net.

\bibitem[{Gruslys et~al.(2016)Gruslys, Munos, Danihelka, Lanctot, and
  Graves}]{DBLP:conf/nips/GruslysMDLG16}
Audrunas Gruslys, R{\'{e}}mi Munos, Ivo Danihelka, Marc Lanctot, and Alex
  Graves. 2016.
\newblock \href
  {https://proceedings.neurips.cc/paper/2016/hash/a501bebf79d570651ff601788ea9d16d-Abstract.html}
  {Memory-efficient backpropagation through time}.
\newblock In \emph{Advances in Neural Information Processing Systems 29: Annual
  Conference on Neural Information Processing Systems 2016, December 5-10,
  2016, Barcelona, Spain}, pages 4125--4133.

\bibitem[{Guo et~al.(2021)Guo, Rush, and Kim}]{DBLP:conf/acl/GuoRK20}
Demi Guo, Alexander~M. Rush, and Yoon Kim. 2021.
\newblock \href {https://doi.org/10.18653/V1/2021.ACL-LONG.378}
  {Parameter-efficient transfer learning with diff pruning}.
\newblock In \emph{Proceedings of the 59th Annual Meeting of the Association
  for Computational Linguistics and the 11th International Joint Conference on
  Natural Language Processing, {ACL/IJCNLP} 2021, (Volume 1: Long Papers),
  Virtual Event, August 1-6, 2021}, pages 4884--4896. Association for
  Computational Linguistics.

\bibitem[{Han et~al.(2024)Han, Gao, Liu, Zhang, and
  Zhang}]{DBLP:journals/corr/abs-2403-14608}
Zeyu Han, Chao Gao, Jinyang Liu, Jeff Zhang, and Sai~Qian Zhang. 2024.
\newblock \href {https://doi.org/10.48550/ARXIV.2403.14608}
  {Parameter-efficient fine-tuning for large models: {A} comprehensive survey}.
\newblock \emph{CoRR}, abs/2403.14608.

\bibitem[{He et~al.(2022{\natexlab{a}})He, Zhou, Ma, Berg{-}Kirkpatrick, and
  Neubig}]{DBLP:conf/iclr/HeZMBN22}
Junxian He, Chunting Zhou, Xuezhe Ma, Taylor Berg{-}Kirkpatrick, and Graham
  Neubig. 2022{\natexlab{a}}.
\newblock \href {https://openreview.net/forum?id=0RDcd5Axok} {Towards a unified
  view of parameter-efficient transfer learning}.
\newblock In \emph{The Tenth International Conference on Learning
  Representations, {ICLR} 2022, Virtual Event, April 25-29, 2022}.
  OpenReview.net.

\bibitem[{He et~al.(2022{\natexlab{b}})He, Zhou, Ma, Berg{-}Kirkpatrick, and
  Neubig}]{he_22}
Junxian He, Chunting Zhou, Xuezhe Ma, Taylor Berg{-}Kirkpatrick, and Graham
  Neubig. 2022{\natexlab{b}}.
\newblock \href {https://openreview.net/forum?id=0RDcd5Axok} {Towards a unified
  view of parameter-efficient transfer learning}.
\newblock In \emph{The Tenth International Conference on Learning
  Representations, {ICLR} 2022, Virtual Event, April 25-29, 2022}.
  OpenReview.net.

\bibitem[{Hendrycks et~al.(2021)Hendrycks, Burns, Basart, Zou, Mazeika, Song,
  and Steinhardt}]{hendrycks_21}
Dan Hendrycks, Collin Burns, Steven Basart, Andy Zou, Mantas Mazeika, Dawn
  Song, and Jacob Steinhardt. 2021.
\newblock \href {https://openreview.net/forum?id=d7KBjmI3GmQ} {Measuring
  massive multitask language understanding}.
\newblock In \emph{9th International Conference on Learning Representations,
  {ICLR} 2021, Virtual Event, Austria, May 3-7, 2021}. OpenReview.net.

\bibitem[{Houlsby et~al.(2019)Houlsby, Giurgiu, Jastrzebski, Morrone,
  de~Laroussilhe, Gesmundo, Attariyan, and
  Gelly}]{DBLP:conf/icml/HoulsbyGJMLGAG19}
Neil Houlsby, Andrei Giurgiu, Stanislaw Jastrzebski, Bruna Morrone, Quentin
  de~Laroussilhe, Andrea Gesmundo, Mona Attariyan, and Sylvain Gelly. 2019.
\newblock \href {http://proceedings.mlr.press/v97/houlsby19a.html}
  {Parameter-efficient transfer learning for {NLP}}.
\newblock In \emph{Proceedings of the 36th International Conference on Machine
  Learning, {ICML} 2019, 9-15 June 2019, Long Beach, California, {USA}},
  volume~97 of \emph{Proceedings of Machine Learning Research}, pages
  2790--2799. {PMLR}.

\bibitem[{Howard and Ruder(2018)}]{ruder_18}
Jeremy Howard and Sebastian Ruder. 2018.
\newblock \href {https://doi.org/10.18653/v1/P18-1031} {Universal language
  model fine-tuning for text classification}.
\newblock In \emph{Proceedings of the 56th Annual Meeting of the Association
  for Computational Linguistics, {ACL} 2018, Melbourne, Australia, July 15-20,
  2018, Volume 1: Long Papers}, pages 328--339. Association for Computational
  Linguistics.

\bibitem[{Hu et~al.(2022)Hu, Shen, Wallis, Allen{-}Zhu, Li, Wang, Wang, and
  Chen}]{hu_22}
Edward~J. Hu, Yelong Shen, Phillip Wallis, Zeyuan Allen{-}Zhu, Yuanzhi Li,
  Shean Wang, Lu~Wang, and Weizhu Chen. 2022.
\newblock \href {https://openreview.net/forum?id=nZeVKeeFYf9} {{LoRA}: Low-rank
  adaptation of large language models}.
\newblock In \emph{The Tenth International Conference on Learning
  Representations, {ICLR} 2022, Virtual Event, April 25-29, 2022}.
  OpenReview.net.

\bibitem[{Jin et~al.(2023)Jin, Zhang, and Zong}]{DBLP:conf/emnlp/JinZZ23}
Feihu Jin, Jiajun Zhang, and Chengqing Zong. 2023.
\newblock \href {https://doi.org/10.18653/V1/2023.EMNLP-MAIN.22}
  {Parameter-efficient tuning for large language model without calculating its
  gradients}.
\newblock In \emph{Proceedings of the 2023 Conference on Empirical Methods in
  Natural Language Processing, {EMNLP} 2023, Singapore, December 6-10, 2023},
  pages 321--330. Association for Computational Linguistics.

\bibitem[{Kurtz et~al.(2020)Kurtz, Kopinsky, Gelashvili, Matveev, Carr, Goin,
  Leiserson, Moore, Shavit, and Alistarh}]{DBLP:conf/icml/KurtzKGMCGLMSA20}
Mark Kurtz, Justin Kopinsky, Rati Gelashvili, Alexander Matveev, John Carr,
  Michael Goin, William~M. Leiserson, Sage Moore, Nir Shavit, and Dan Alistarh.
  2020.
\newblock \href {http://proceedings.mlr.press/v119/kurtz20a.html} {Inducing and
  exploiting activation sparsity for fast inference on deep neural networks}.
\newblock In \emph{Proceedings of the 37th International Conference on Machine
  Learning, {ICML} 2020, 13-18 July 2020, Virtual Event}, volume 119 of
  \emph{Proceedings of Machine Learning Research}, pages 5533--5543. {PMLR}.

\bibitem[{Lawton et~al.(2023)Lawton, Kumar, Thattai, Galstyan, and
  Steeg}]{DBLP:conf/acl/LawtonKTGS23}
Neal Lawton, Anoop Kumar, Govind Thattai, Aram Galstyan, and Greg~Ver Steeg.
  2023.
\newblock \href {https://doi.org/10.18653/V1/2023.FINDINGS-ACL.539} {Neural
  architecture search for parameter-efficient fine-tuning of large pre-trained
  language models}.
\newblock In \emph{Findings of the Association for Computational Linguistics:
  {ACL} 2023, Toronto, Canada, July 9-14, 2023}, pages 8506--8515. Association
  for Computational Linguistics.

\bibitem[{Lee et~al.(2023)Lee, Hunter, and Ruiz}]{lee_23}
Ariel~N. Lee, Cole~J. Hunter, and Nataniel Ruiz. 2023.
\newblock \href {https://doi.org/10.48550/arXiv.2308.07317} {Platypus: Quick,
  cheap, and powerful refinement of llms}.
\newblock In \emph{NeurIPS 2023 Workshop on Instruction Tuning and Instruction
  Following}.

\bibitem[{Lee et~al.(2019)Lee, Tang, and Lin}]{lee_19}
Jaejun Lee, Raphael Tang, and Jimmy Lin. 2019.
\newblock \href {https://arxiv.org/abs/1911.03090} {What would elsa do?
  freezing layers during transformer fine-tuning}.
\newblock \emph{CoRR}, abs/1911.03090.

\bibitem[{Li and Zhang(2021)}]{DBLP:conf/nips/LiZ21}
Dongyue Li and Hongyang~R. Zhang. 2021.
\newblock \href
  {https://proceedings.neurips.cc/paper/2021/hash/e4a93f0332b2519177ed55741ea4e5e7-Abstract.html}
  {Improved regularization and robustness for fine-tuning in neural networks}.
\newblock In \emph{Advances in Neural Information Processing Systems 34: Annual
  Conference on Neural Information Processing Systems 2021, NeurIPS 2021,
  December 6-14, 2021, virtual}, pages 27249--27262.

\bibitem[{Li and Liang(2021)}]{DBLP:conf/acl/LiL20}
Xiang~Lisa Li and Percy Liang. 2021.
\newblock \href {https://doi.org/10.18653/V1/2021.ACL-LONG.353} {Prefix-tuning:
  Optimizing continuous prompts for generation}.
\newblock In \emph{Proceedings of the 59th Annual Meeting of the Association
  for Computational Linguistics and the 11th International Joint Conference on
  Natural Language Processing, {ACL/IJCNLP} 2021, (Volume 1: Long Papers),
  Virtual Event, August 1-6, 2021}, pages 4582--4597. Association for
  Computational Linguistics.

\bibitem[{Li et~al.(2022)Li, Tram{\`{e}}r, Liang, and
  Hashimoto}]{DBLP:conf/iclr/LiTLH22}
Xuechen Li, Florian Tram{\`{e}}r, Percy Liang, and Tatsunori Hashimoto. 2022.
\newblock \href {https://openreview.net/forum?id=bVuP3ltATMz} {Large language
  models can be strong differentially private learners}.
\newblock In \emph{The Tenth International Conference on Learning
  Representations, {ICLR} 2022, Virtual Event, April 25-29, 2022}.
  OpenReview.net.

\bibitem[{Liao et~al.(2023)Liao, Tan, and Monz}]{DBLP:conf/nips/LiaoTM23}
Baohao Liao, Shaomu Tan, and Christof Monz. 2023.
\newblock \href
  {http://papers.nips.cc/paper\_files/paper/2023/hash/3151e460c41ba67dc55412861184ef35-Abstract-Conference.html}
  {Make pre-trained model reversible: From parameter to memory efficient
  fine-tuning}.
\newblock In \emph{Advances in Neural Information Processing Systems 36: Annual
  Conference on Neural Information Processing Systems 2023, NeurIPS 2023, New
  Orleans, LA, USA, December 10 - 16, 2023}.

\bibitem[{Lin et~al.(2022)Lin, Hilton, and Evans}]{lin_22}
Stephanie Lin, Jacob Hilton, and Owain Evans. 2022.
\newblock \href {https://doi.org/10.18653/v1/2022.acl-long.229} {Truthfulqa:
  Measuring how models mimic human falsehoods}.
\newblock In \emph{Proceedings of the 60th Annual Meeting of the Association
  for Computational Linguistics (Volume 1: Long Papers), {ACL} 2022, Dublin,
  Ireland, May 22-27, 2022}, pages 3214--3252. Association for Computational
  Linguistics.

\bibitem[{Lin et~al.(2024)Lin, Ma, Chu, Jin, Yang, Wang, and
  Mei}]{DBLP:journals/corr/abs-2404-09610}
Yang Lin, Xinyu Ma, Xu~Chu, Yujie Jin, Zhibang Yang, Yasha Wang, and Hong Mei.
  2024.
\newblock \href {https://doi.org/10.48550/ARXIV.2404.09610} {{LoRA} dropout as
  a sparsity regularizer for overfitting control}.
\newblock \emph{CoRR}, abs/2404.09610.

\bibitem[{Liu et~al.(2022)Liu, Ji, Fu, Tam, Du, Yang, and Tang}]{liu_21}
Xiao Liu, Kaixuan Ji, Yicheng Fu, Weng Tam, Zhengxiao Du, Zhilin Yang, and Jie
  Tang. 2022.
\newblock \href {https://doi.org/10.18653/v1/2022.acl-short.8} {{P}-tuning:
  Prompt tuning can be comparable to fine-tuning across scales and tasks}.
\newblock In \emph{Proceedings of the 60th Annual Meeting of the Association
  for Computational Linguistics (Volume 2: Short Papers)}, pages 61--68,
  Dublin, Ireland. Association for Computational Linguistics.

\bibitem[{Liu et~al.(2024)Liu, Oguz, Zhao, Chang, Stock, Mehdad, Shi,
  Krishnamoorthi, and Chandra}]{liu_23}
Zechun Liu, Barlas Oguz, Changsheng Zhao, Ernie Chang, Pierre Stock, Yashar
  Mehdad, Yangyang Shi, Raghuraman Krishnamoorthi, and Vikas Chandra. 2024.
\newblock \href {https://aclanthology.org/2024.findings-acl.26} {{LLM-QAT:}
  data-free quantization aware training for large language models}.
\newblock In \emph{Findings of the Association for Computational Linguistics,
  {ACL} 2024, Bangkok, Thailand and virtual meeting, August 11-16, 2024}, pages
  467--484. Association for Computational Linguistics.

\bibitem[{Liu et~al.(2023)Liu, Wang, Dao, Zhou, Yuan, Song, Shrivastava, Zhang,
  Tian, R{\'{e}}, and Chen}]{DBLP:conf/icml/LiuWDZY0S0TRC23}
Zichang Liu, Jue Wang, Tri Dao, Tianyi Zhou, Binhang Yuan, Zhao Song, Anshumali
  Shrivastava, Ce~Zhang, Yuandong Tian, Christopher R{\'{e}}, and Beidi Chen.
  2023.
\newblock \href {https://proceedings.mlr.press/v202/liu23am.html} {Deja vu:
  Contextual sparsity for efficient llms at inference time}.
\newblock In \emph{International Conference on Machine Learning, {ICML} 2023,
  23-29 July 2023, Honolulu, Hawaii, {USA}}, volume 202 of \emph{Proceedings of
  Machine Learning Research}, pages 22137--22176. {PMLR}.

\bibitem[{Malladi et~al.(2023)Malladi, Gao, Nichani, Damian, Lee, Chen, and
  Arora}]{DBLP:conf/nips/MalladiGNDL0A23}
Sadhika Malladi, Tianyu Gao, Eshaan Nichani, Alex Damian, Jason~D. Lee, Danqi
  Chen, and Sanjeev Arora. 2023.
\newblock \href
  {http://papers.nips.cc/paper\_files/paper/2023/hash/a627810151be4d13f907ac898ff7e948-Abstract-Conference.html}
  {Fine-tuning language models with just forward passes}.
\newblock In \emph{Advances in Neural Information Processing Systems 36: Annual
  Conference on Neural Information Processing Systems 2023, NeurIPS 2023, New
  Orleans, LA, USA, December 10 - 16, 2023}.

\bibitem[{Mao et~al.(2022)Mao, Mathias, Hou, Almahairi, Ma, Han, Yih, and
  Khabsa}]{DBLP:conf/acl/MaoMHAM0YK22}
Yuning Mao, Lambert Mathias, Rui Hou, Amjad Almahairi, Hao Ma, Jiawei Han,
  Scott Yih, and Madian Khabsa. 2022.
\newblock \href {https://doi.org/10.18653/V1/2022.ACL-LONG.433} {{UniPELT}: {A}
  unified framework for parameter-efficient language model tuning}.
\newblock In \emph{Proceedings of the 60th Annual Meeting of the Association
  for Computational Linguistics (Volume 1: Long Papers), {ACL} 2022, Dublin,
  Ireland, May 22-27, 2022}, pages 6253--6264. Association for Computational
  Linguistics.

\bibitem[{Mishra et~al.(2022)Mishra, Khashabi, Baral, and
  Hajishirzi}]{DBLP:conf/acl/MishraKBH22}
Swaroop Mishra, Daniel Khashabi, Chitta Baral, and Hannaneh Hajishirzi. 2022.
\newblock \href {https://doi.org/10.18653/V1/2022.ACL-LONG.244} {Cross-task
  generalization via natural language crowdsourcing instructions}.
\newblock In \emph{Proceedings of the 60th Annual Meeting of the Association
  for Computational Linguistics (Volume 1: Long Papers), {ACL} 2022, Dublin,
  Ireland, May 22-27, 2022}, pages 3470--3487. Association for Computational
  Linguistics.

\bibitem[{Mukherjee et~al.(2023)Mukherjee, Mitra, Jawahar, Agarwal, Palangi,
  and Awadallah}]{mukherjee_23}
Subhabrata Mukherjee, Arindam Mitra, Ganesh Jawahar, Sahaj Agarwal, Hamid
  Palangi, and Ahmed~Hassan Awadallah. 2023.
\newblock \href {https://doi.org/10.48550/arXiv.2306.02707} {Orca: Progressive
  learning from complex explanation traces of {GPT-4}}.
\newblock \emph{CoRR}, abs/2306.02707.

\bibitem[{OpenAI(2023)}]{DBLP:journals/corr/abs-2303-08774}
OpenAI. 2023.
\newblock \href {https://doi.org/10.48550/ARXIV.2303.08774} {{GPT-4} technical
  report}.
\newblock \emph{CoRR}, abs/2303.08774.

\bibitem[{Ouyang et~al.(2022)Ouyang, Wu, Jiang, Almeida, Wainwright, Mishkin,
  Zhang, Agarwal, Slama, Ray, Schulman, Hilton, Kelton, Miller, Simens, Askell,
  Welinder, Christiano, Leike, and Lowe}]{ouyang_22}
Long Ouyang, Jeffrey Wu, Xu~Jiang, Diogo Almeida, Carroll~L. Wainwright, Pamela
  Mishkin, Chong Zhang, Sandhini Agarwal, Katarina Slama, Alex Ray, John
  Schulman, Jacob Hilton, Fraser Kelton, Luke Miller, Maddie Simens, Amanda
  Askell, Peter Welinder, Paul~F. Christiano, Jan Leike, and Ryan Lowe. 2022.
\newblock \href
  {http://papers.nips.cc/paper\_files/paper/2022/hash/b1efde53be364a73914f58805a001731-Abstract-Conference.html}
  {Training language models to follow instructions with human feedback}.
\newblock In \emph{NeurIPS}.

\bibitem[{Pfeiffer et~al.(2021)Pfeiffer, Kamath, R{\"{u}}ckl{\'{e}}, Cho, and
  Gurevych}]{DBLP:conf/eacl/PfeifferKRCG21}
Jonas Pfeiffer, Aishwarya Kamath, Andreas R{\"{u}}ckl{\'{e}}, Kyunghyun Cho,
  and Iryna Gurevych. 2021.
\newblock \href {https://doi.org/10.18653/V1/2021.EACL-MAIN.39} {Adapterfusion:
  Non-destructive task composition for transfer learning}.
\newblock In \emph{Proceedings of the 16th Conference of the European Chapter
  of the Association for Computational Linguistics: Main Volume, {EACL} 2021,
  Online, April 19 - 23, 2021}, pages 487--503. Association for Computational
  Linguistics.

\bibitem[{Pfeiffer et~al.(2023)Pfeiffer, Ruder, Vulic, and Ponti}]{pfeiffer_23}
Jonas Pfeiffer, Sebastian Ruder, Ivan Vulic, and Edoardo~M. Ponti. 2023.
\newblock \href {https://openreview.net/forum?id=z9EkXfvxta} {Modular deep
  learning}.
\newblock \emph{Trans. Mach. Learn. Res.}, 2023.

\bibitem[{Qin and Eisner(2021)}]{qin_21}
Guanghui Qin and Jason Eisner. 2021.
\newblock \href {https://doi.org/10.18653/v1/2021.naacl-main.410} {Learning how
  to ask: Querying lms with mixtures of soft prompts}.
\newblock In \emph{Proceedings of the 2021 Conference of the North American
  Chapter of the Association for Computational Linguistics: Human Language
  Technologies, {NAACL-HLT} 2021, Online, June 6-11, 2021}, pages 5203--5212.
  Association for Computational Linguistics.

\bibitem[{Raffel et~al.(2020)Raffel, Shazeer, Roberts, Lee, Narang, Matena,
  Zhou, Li, and Liu}]{raffel_20}
Colin Raffel, Noam Shazeer, Adam Roberts, Katherine Lee, Sharan Narang, Michael
  Matena, Yanqi Zhou, Wei Li, and Peter~J. Liu. 2020.
\newblock \href {http://jmlr.org/papers/v21/20-074.html} {Exploring the limits
  of transfer learning with a unified text-to-text transformer}.
\newblock \emph{J. Mach. Learn. Res.}, 21:140:1--140:67.

\bibitem[{Sakaguchi et~al.(2020)Sakaguchi, Bras, Bhagavatula, and
  Choi}]{DBLP:conf/aaai/SakaguchiBBC20}
Keisuke Sakaguchi, Ronan~Le Bras, Chandra Bhagavatula, and Yejin Choi. 2020.
\newblock \href {https://doi.org/10.1609/AAAI.V34I05.6399} {Winogrande: An
  adversarial winograd schema challenge at scale}.
\newblock In \emph{The Thirty-Fourth {AAAI} Conference on Artificial
  Intelligence, {AAAI} 2020, The Thirty-Second Innovative Applications of
  Artificial Intelligence Conference, {IAAI} 2020, The Tenth {AAAI} Symposium
  on Educational Advances in Artificial Intelligence, {EAAI} 2020, New York,
  NY, USA, February 7-12, 2020}, pages 8732--8740. {AAAI} Press.

\bibitem[{Schick and Sch{\"{u}}tze(2021)}]{DBLP:conf/naacl/SchickS21}
Timo Schick and Hinrich Sch{\"{u}}tze. 2021.
\newblock \href {https://doi.org/10.18653/v1/2021.naacl-main.185} {It's not
  just size that matters: Small language models are also few-shot learners}.
\newblock In \emph{Proceedings of the 2021 Conference of the North American
  Chapter of the Association for Computational Linguistics: Human Language
  Technologies, {NAACL-HLT} 2021, Online, June 6-11, 2021}, pages 2339--2352.
  Association for Computational Linguistics.

\bibitem[{Shazeer and Stern(2018)}]{DBLP:conf/icml/ShazeerS18}
Noam Shazeer and Mitchell Stern. 2018.
\newblock \href {http://proceedings.mlr.press/v80/shazeer18a.html} {Adafactor:
  Adaptive learning rates with sublinear memory cost}.
\newblock In \emph{Proceedings of the 35th International Conference on Machine
  Learning, {ICML} 2018, Stockholmsm{\"{a}}ssan, Stockholm, Sweden, July 10-15,
  2018}, volume~80 of \emph{Proceedings of Machine Learning Research}, pages
  4603--4611. {PMLR}.

\bibitem[{Simoulin et~al.(2023)Simoulin, Park, Liu, and
  Yang}]{simoulin2023memoryefficient}
Antoine Simoulin, Namyong Park, Xiaoyi Liu, and Grey Yang. 2023.
\newblock \href {https://openreview.net/forum?id=zaNbLceVwm} {Memory-efficient
  selective fine-tuning}.
\newblock In \emph{Workshop on Efficient Systems for Foundation Models @
  ICML2023}.

\bibitem[{Spall(1992)}]{Spall1992MultivariateSA}
James~C. Spall. 1992.
\newblock \href {https://api.semanticscholar.org/CorpusID:122365276}
  {Multivariate stochastic approximation using a simultaneous perturbation
  gradient approximation}.
\newblock \emph{IEEE Transactions on Automatic Control}, 37:332--341.

\bibitem[{Sung et~al.(2022)Sung, Cho, and Bansal}]{sung2022lst}
Yi{-}Lin Sung, Jaemin Cho, and Mohit Bansal. 2022.
\newblock \href {https://openreview.net/forum?id=isPnnaTZaP5} {{LST}: Ladder
  side-tuning for parameter and memory efficient transfer learning}.
\newblock In \emph{NeurIPS}.

\bibitem[{Sung et~al.(2021)Sung, Nair, and Raffel}]{DBLP:conf/nips/SungNR21}
Yi{-}Lin Sung, Varun Nair, and Colin Raffel. 2021.
\newblock \href
  {https://proceedings.neurips.cc/paper/2021/hash/cb2653f548f8709598e8b5156738cc51-Abstract.html}
  {Training neural networks with fixed sparse masks}.
\newblock In \emph{Advances in Neural Information Processing Systems 34: Annual
  Conference on Neural Information Processing Systems 2021, NeurIPS 2021,
  December 6-14, 2021, virtual}, pages 24193--24205.

\bibitem[{Taori et~al.(2023)Taori, Gulrajani, Zhang, Dubois, Li, Guestrin,
  Liang, and Hashimoto}]{alpaca}
Rohan Taori, Ishaan Gulrajani, Tianyi Zhang, Yann Dubois, Xuechen Li, Carlos
  Guestrin, Percy Liang, and Tatsunori~B. Hashimoto. 2023.
\newblock Stanford alpaca: An instruction-following llama model.
\newblock \url{https://github.com/tatsu-lab/stanford_alpaca}.

\bibitem[{Touvron et~al.(2023)Touvron, Martin, Stone, Albert, Almahairi,
  Babaei, Bashlykov, Batra, Bhargava, Bhosale, Bikel, Blecher, Canton{-}Ferrer,
  Chen, Cucurull, Esiobu, Fernandes, Fu, Fu, Fuller, Gao, Goswami, Goyal,
  Hartshorn, Hosseini, Hou, Inan, Kardas, Kerkez, Khabsa, Kloumann, Korenev,
  Koura, Lachaux, Lavril, Lee, Liskovich, Lu, Mao, Martinet, Mihaylov, Mishra,
  Molybog, Nie, Poulton, Reizenstein, Rungta, Saladi, Schelten, Silva, Smith,
  Subramanian, Tan, Tang, Taylor, Williams, Kuan, Xu, Yan, Zarov, Zhang, Fan,
  Kambadur, Narang, Rodriguez, Stojnic, Edunov, and Scialom}]{touvron_23}
Hugo Touvron, Louis Martin, Kevin Stone, Peter Albert, Amjad Almahairi, Yasmine
  Babaei, Nikolay Bashlykov, Soumya Batra, Prajjwal Bhargava, Shruti Bhosale,
  Dan Bikel, Lukas Blecher, Cristian Canton{-}Ferrer, Moya Chen, Guillem
  Cucurull, David Esiobu, Jude Fernandes, Jeremy Fu, Wenyin Fu, Brian Fuller,
  Cynthia Gao, Vedanuj Goswami, Naman Goyal, Anthony Hartshorn, Saghar
  Hosseini, Rui Hou, Hakan Inan, Marcin Kardas, Viktor Kerkez, Madian Khabsa,
  Isabel Kloumann, Artem Korenev, Punit~Singh Koura, Marie{-}Anne Lachaux,
  Thibaut Lavril, Jenya Lee, Diana Liskovich, Yinghai Lu, Yuning Mao, Xavier
  Martinet, Todor Mihaylov, Pushkar Mishra, Igor Molybog, Yixin Nie, Andrew
  Poulton, Jeremy Reizenstein, Rashi Rungta, Kalyan Saladi, Alan Schelten, Ruan
  Silva, Eric~Michael Smith, Ranjan Subramanian, Xiaoqing~Ellen Tan, Binh Tang,
  Ross Taylor, Adina Williams, Jian~Xiang Kuan, Puxin Xu, Zheng Yan, Iliyan
  Zarov, Yuchen Zhang, Angela Fan, Melanie Kambadur, Sharan Narang,
  Aur{\'{e}}lien Rodriguez, Robert Stojnic, Sergey Edunov, and Thomas Scialom.
  2023.
\newblock \href {https://doi.org/10.48550/arXiv.2307.09288} {Llama 2: Open
  foundation and fine-tuned chat models}.
\newblock \emph{CoRR}, abs/2307.09288.

\bibitem[{Valipour et~al.(2023)Valipour, Rezagholizadeh, Kobyzev, and
  Ghodsi}]{DBLP:conf/eacl/ValipourRKG23}
Mojtaba Valipour, Mehdi Rezagholizadeh, Ivan Kobyzev, and Ali Ghodsi. 2023.
\newblock \href {https://doi.org/10.18653/V1/2023.EACL-MAIN.239} {{DyLoRA}:
  Parameter-efficient tuning of pre-trained models using dynamic search-free
  low-rank adaptation}.
\newblock In \emph{Proceedings of the 17th Conference of the European Chapter
  of the Association for Computational Linguistics, {EACL} 2023, Dubrovnik,
  Croatia, May 2-6, 2023}, pages 3266--3279. Association for Computational
  Linguistics.

\bibitem[{Wang et~al.(2018)Wang, Singh, Michael, Hill, Levy, and
  Bowman}]{wang_18}
Alex Wang, Amanpreet Singh, Julian Michael, Felix Hill, Omer Levy, and Samuel
  Bowman. 2018.
\newblock \href {https://doi.org/10.18653/v1/W18-5446} {{GLUE}: A multi-task
  benchmark and analysis platform for natural language understanding}.
\newblock In \emph{Proceedings of the 2018 {EMNLP} Workshop {B}lackbox{NLP}:
  Analyzing and Interpreting Neural Networks for {NLP}}, pages 353--355,
  Brussels, Belgium. Association for Computational Linguistics.

\bibitem[{Wang et~al.(2023)Wang, Kordi, Mishra, Liu, Smith, Khashabi, and
  Hajishirzi}]{wang_23}
Yizhong Wang, Yeganeh Kordi, Swaroop Mishra, Alisa Liu, Noah~A. Smith, Daniel
  Khashabi, and Hannaneh Hajishirzi. 2023.
\newblock \href {https://doi.org/10.18653/v1/2023.acl-long.754} {Self-instruct:
  Aligning language models with self-generated instructions}.
\newblock In \emph{Proceedings of the 61st Annual Meeting of the Association
  for Computational Linguistics (Volume 1: Long Papers), {ACL} 2023, Toronto,
  Canada, July 9-14, 2023}, pages 13484--13508. Association for Computational
  Linguistics.

\bibitem[{Wei et~al.(2022)Wei, Bosma, Zhao, Guu, Yu, Lester, Du, Dai, and
  Le}]{wei_22}
Jason Wei, Maarten Bosma, Vincent~Y. Zhao, Kelvin Guu, Adams~Wei Yu, Brian
  Lester, Nan Du, Andrew~M. Dai, and Quoc~V. Le. 2022.
\newblock \href {https://openreview.net/forum?id=gEZrGCozdqR} {Finetuned
  language models are zero-shot learners}.
\newblock In \emph{The Tenth International Conference on Learning
  Representations, {ICLR} 2022, Virtual Event, April 25-29, 2022}.
  OpenReview.net.

\bibitem[{Xu et~al.(2023)Xu, Xie, Qin, Tao, and
  Wang}]{DBLP:journals/corr/abs-2312-12148}
Lingling Xu, Haoran Xie, Si{-}Zhao~Joe Qin, Xiaohui Tao, and Fu~Lee Wang. 2023.
\newblock \href {https://doi.org/10.48550/ARXIV.2312.12148}
  {Parameter-efficient fine-tuning methods for pretrained language models: {A}
  critical review and assessment}.
\newblock \emph{CoRR}, abs/2312.12148.

\bibitem[{Yang et~al.(2024)Yang, Robeyns, Wang, and
  Aitchison}]{DBLP:conf/iclr/YangRWA24}
Adam~X. Yang, Maxime Robeyns, Xi~Wang, and Laurence Aitchison. 2024.
\newblock \href {https://openreview.net/forum?id=FJiUyzOF1m} {Bayesian low-rank
  adaptation for large language models}.
\newblock In \emph{The Twelfth International Conference on Learning
  Representations, {ICLR} 2024, Vienna, Austria, May 7-11, 2024}.
  OpenReview.net.

\bibitem[{Zaken et~al.(2022)Zaken, Goldberg, and Ravfogel}]{zaken_22}
Elad~Ben Zaken, Yoav Goldberg, and Shauli Ravfogel. 2022.
\newblock \href {https://doi.org/10.18653/v1/2022.acl-short.1} {Bitfit: Simple
  parameter-efficient fine-tuning for transformer-based masked
  language-models}.
\newblock In \emph{Proceedings of the 60th Annual Meeting of the Association
  for Computational Linguistics (Volume 2: Short Papers), {ACL} 2022, Dublin,
  Ireland, May 22-27, 2022}, pages 1--9. Association for Computational
  Linguistics.

\bibitem[{Zellers et~al.(2019)Zellers, Holtzman, Bisk, Farhadi, and
  Choi}]{zellers19}
Rowan Zellers, Ari Holtzman, Yonatan Bisk, Ali Farhadi, and Yejin Choi. 2019.
\newblock \href {https://doi.org/10.18653/v1/p19-1472} {Hellaswag: Can a
  machine really finish your sentence?}
\newblock In \emph{Proceedings of the 57th Conference of the Association for
  Computational Linguistics, {ACL} 2019, Florence, Italy, July 28- August 2,
  2019, Volume 1: Long Papers}, pages 4791--4800. Association for Computational
  Linguistics.

\bibitem[{Zhang et~al.(2020)Zhang, Sax, Zamir, Guibas, and Malik}]{zhang_20}
Jeffrey~O. Zhang, Alexander Sax, Amir Zamir, Leonidas~J. Guibas, and Jitendra
  Malik. 2020.
\newblock \href {https://doi.org/10.1007/978-3-030-58580-8\_41} {Side-tuning:
  {A} baseline for network adaptation via additive side networks}.
\newblock In \emph{Computer Vision - {ECCV} 2020 - 16th European Conference,
  Glasgow, UK, August 23-28, 2020, Proceedings, Part {III}}, volume 12348 of
  \emph{Lecture Notes in Computer Science}, pages 698--714. Springer.

\bibitem[{Zhang et~al.(2023{\natexlab{a}})Zhang, Zhang, Shi, Chu, and
  Li}]{DBLP:journals/corr/abs-2308-03303}
Longteng Zhang, Lin Zhang, Shaohuai Shi, Xiaowen Chu, and Bo~Li.
  2023{\natexlab{a}}.
\newblock \href {https://doi.org/10.48550/ARXIV.2308.03303} {{LoRA-FA}:
  Memory-efficient low-rank adaptation for large language models fine-tuning}.
\newblock \emph{CoRR}, abs/2308.03303.

\bibitem[{Zhang et~al.(2023{\natexlab{b}})Zhang, Chen, Bukharin, He, Cheng,
  Chen, and Zhao}]{DBLP:conf/iclr/ZhangCBH0CZ23}
Qingru Zhang, Minshuo Chen, Alexander Bukharin, Pengcheng He, Yu~Cheng, Weizhu
  Chen, and Tuo Zhao. 2023{\natexlab{b}}.
\newblock \href {https://openreview.net/pdf?id=lq62uWRJjiY} {Adaptive budget
  allocation for parameter-efficient fine-tuning}.
\newblock In \emph{The Eleventh International Conference on Learning
  Representations, {ICLR} 2023, Kigali, Rwanda, May 1-5, 2023}. OpenReview.net.

\bibitem[{Zhang et~al.(2022{\natexlab{a}})Zhang, Roller, Goyal, Artetxe, Chen,
  Chen, Dewan, Diab, Li, Lin, Mihaylov, Ott, Shleifer, Shuster, Simig, Koura,
  Sridhar, Wang, and Zettlemoyer}]{zhang_22}
Susan Zhang, Stephen Roller, Naman Goyal, Mikel Artetxe, Moya Chen, Shuohui
  Chen, Christopher Dewan, Mona~T. Diab, Xian Li, Xi~Victoria Lin, Todor
  Mihaylov, Myle Ott, Sam Shleifer, Kurt Shuster, Daniel Simig, Punit~Singh
  Koura, Anjali Sridhar, Tianlu Wang, and Luke Zettlemoyer. 2022{\natexlab{a}}.
\newblock \href {https://doi.org/10.48550/arXiv.2205.01068} {{OPT:} open
  pre-trained transformer language models}.
\newblock \emph{CoRR}, abs/2205.01068.

\bibitem[{Zhang et~al.(2022{\natexlab{b}})Zhang, Zhou, and
  Liu}]{DBLP:journals/corr/abs-2206-04673}
Yuanhan Zhang, Kaiyang Zhou, and Ziwei Liu. 2022{\natexlab{b}}.
\newblock \href {https://doi.org/10.48550/ARXIV.2206.04673} {Neural prompt
  search}.
\newblock \emph{CoRR}, abs/2206.04673.

\bibitem[{Zhao et~al.(2024)Zhao, Zhang, Chen, Wang, Anandkumar, and
  Tian}]{zhao2024galore}
Jiawei Zhao, Zhenyu Zhang, Beidi Chen, Zhangyang Wang, Anima Anandkumar, and
  Yuandong Tian. 2024.
\newblock \href {https://openreview.net/forum?id=hYHsrKDiX7} {{GaLore}:
  Memory-efficient {LLM} training by gradient low-rank projection}.
\newblock In \emph{Forty-first International Conference on Machine Learning,
  {ICML} 2024}.

\bibitem[{Zhou et~al.(2024)Zhou, Wan, Vulic, and
  Korhonen}]{DBLP:journals/tacl/ZhouWVK24}
Han Zhou, Xingchen Wan, Ivan Vulic, and Anna Korhonen. 2024.
\newblock \href {https://doi.org/10.1162/TACL\_A\_00662} {{AutoPEFT}: Automatic
  configuration search for parameter-efficient fine-tuning}.
\newblock \emph{Trans. Assoc. Comput. Linguistics}, 12:525--542.

\bibitem[{Zhu et~al.(2021)Zhu, Feng, Zhao, Wang, and
  Li}]{DBLP:conf/emnlp/ZhuFZWL21}
Yaoming Zhu, Jiangtao Feng, Chengqi Zhao, Mingxuan Wang, and Lei Li. 2021.
\newblock \href {https://doi.org/10.18653/V1/2021.FINDINGS-EMNLP.240}
  {Counter-interference adapter for multilingual machine translation}.
\newblock In \emph{Findings of the Association for Computational Linguistics:
  {EMNLP} 2021, Virtual Event / Punta Cana, Dominican Republic, 16-20 November,
  2021}, pages 2812--2823. Association for Computational Linguistics.

\end{thebibliography}
\end{document}